\documentclass{ecai}

\usepackage{assets/my_settings}

\def\eqref#1{equation~\ref{#1}}

\def\1{\bm{1}}

\def\vtheta{{\bm{\theta}}}

\def\vx{{\bm{x}}}

\def\mX{{\bm{X}}}

\DeclareMathAlphabet{\mathsfit}{\encodingdefault}{\sfdefault}{m}{sl}
\SetMathAlphabet{\mathsfit}{bold}{\encodingdefault}{\sfdefault}{bx}{n}

\def\sX{{\mathbb{X}}}

\usepackage{xr}
\usepackage{hyperref} 

\begin{document}


\begin{frontmatter}


\paperid{1856} 


\title{Unveiling the Power of Sparse Neural Networks for Feature Selection}


\author[A]{\fnms{Zahra}~\snm{Atashgahi}\thanks{Corresponding Author. Email: z.atashgahi@utwente.nl.\\ Accepted for publication at ECAI 2024 }}
\author[B]{\fnms{Tennison}~\snm{Liu}}
\author[C]{\fnms{Mykola}~\snm{Pechenizkiy}}
\author[A]{\fnms{Raymond}~\snm{Veldhuis}}
\author[D,C]{\fnms{Decebal Constantin}~\snm{Mocanu}}
\author[B]{\fnms{Mihaela}~\snm{van der Schaar}}
	
\address[A]{Faculty of Electrical Engineering, Mathematics and Computer Science, University of Twente, The Netherlands}\address[B]{Department of Applied Mathematics and Theoretical Physics, University of Cambridge, United Kingdom}
\address[C]{Department of Mathematics and Computer Science, Eindhoven University of Technology, The Netherlands}
\address[D]{Department of Computer Science, University of Luxembourg, Luxembourg}



\begin{abstract}

Sparse Neural Networks (SNNs) have emerged as powerful tools for efficient feature selection. Leveraging the dynamic sparse training (DST) algorithms within SNNs has demonstrated promising feature selection capabilities while drastically reducing computational overheads. Despite these advancements, several critical aspects remain insufficiently explored for feature selection. Questions persist regarding the choice of the DST algorithm for network training, the choice of metric for ranking features/neurons, and the comparative performance of these methods across diverse datasets when compared to dense networks. This paper addresses these gaps by presenting a comprehensive systematic analysis of feature selection with sparse neural networks. Moreover, we introduce a novel metric considering sparse neural network characteristics, which is designed to quantify feature importance within the context of SNNs. Our findings show that feature selection with SNNs trained with DST algorithms can achieve, on average, more than $50\%$ memory and $55\%$ FLOPs reduction compared to the dense networks, while outperforming them in terms of the quality of the selected features. Our code and the supplementary material are available on GitHub (\url{https://github.com/zahraatashgahi/Neuron-Attribution}).
\end{abstract}

\end{frontmatter}

\section{Introduction}\label{sec:introduction}

\looseness=-1
With the ever-increasing generation of big data, high-dimensional data has become ubiquitous in various fields such as health care, bioinformatics, and social media. Yet, high dimensionality poses substantial challenges for the analysis and interpretation of data such as, curse of dimensionality, overfitting, and high memory and computation demands \cite{li2017feature}. 

\looseness=-1
Feature selection emerges as a pivotal approach to address the challenges raised by high-dimensional data. It selects the most relevant attributes of the data for the final learning task \cite{chandrashekar2014survey}. Feature selection can reduce the computational, memory, and as a result energy costs, increase interpretability, decrease data collection costs, and potentially enhance the model generalization. 

\looseness=-1
Lately, neural networks have emerged as a powerful tool for feature selection. Deep learning methods gain increasing attention in various tasks due to their intrinsic attributes: automatic feature engineering, learning from data streams, facilitation of multi-source learning (multi-modal/multi-view learning), parameters pre-training, and the feasibility of an end-to-end data processing paradigm. This made them attractive for learning feature representations and AutoML \cite{arik2021tabnet, somepalli2021saint, gharibshah2022local, borisov2022deep, he2021automl}. One intriguing advantage of neural network feature selection compared to most traditional approaches is the ability to capture complex non-linear dependencies among features. Notably, neural network-based feature selection has exhibited substantial success, often outperforming conventional feature selection methods in identifying more relevant features for the final prediction tasks \cite{balin2019concrete, imrie2022composite, lemhadri2021lassonet, yamada2020feature, jia2023effective, singh2023fsnet}. Inspired by Lasso \cite{tibshirani1996regression}, a subset of neural network-based feature selection methods introduces sparsity within networks to perform feature selection. \cite{yamada2020feature} introduces sparsity in the input features of a neural network via stochastic gates. \cite{lemhadri2021lassonet} sparsifies the network by selecting the relevant input features for the entire network to perform global feature selection. 

\looseness=-1
However, a main challenge with the existing neural network feature selection method is the high computational cost due to their large over-parameterized networks, particularly when applied to high-dimensional data \cite{atashgahi2022quick}. This makes the deployment of such models infeasible in low-resource environments, e.g., mobile phones. In addition, in extreme cases, training and deploying over parameterized deep learning models, significantly raise energy consumption in data centers, resulting in high carbon emissions exceeding a human's annual carbon footprint \cite{strubell2020energy}. 

\looseness=-1
Therefore, another group of works exploits the characteristics of a sparse neural network (SNN) \cite{hoefler2021sparsity} trained with dynamic sparse training (DST) \cite{mocanu2018scalable} to find the most important attributes of a dataset. This line of works first initiated by \cite{atashgahi2022quick} and followed by \cite{sokar2022pay, atashgahi2022supervised} has demonstrated competitive feature selection capabilities, comparable to state-of-the-art algorithms, all while maintaining computational efficiency. Unlike the methods that regularize weights to achieve sparsity, the latter group of works exploits hard sparsity (zero-out weights). Moreover, they exploit dynamic sparse training to train the network sparsely from scratch to be efficient during the entire training process. Besides, they deploy sparsity in all layers and not only the initial layer; therefore, they are much more computationally efficient.

\looseness=-1
Despite showcasing superior feature selection performance coupled with increased efficiency, certain ambiguities persist concerning feature selection with sparse neural networks. \textit{Is sparsity beneficial for all datasets when performing feature selection? What metric to choose for each dataset when measuring the feature's importance? How does the choice of the DST algorithm affect the feature selection performance?}

\looseness=-1
In this paper, we provide an in-depth analysis of supervised feature selection with SNNs trained with dynamic sparse training. Our contributions are:

\begin{itemize}
    \looseness=-1
    \item We conduct a novel and extensive exploration and analysis of feature selection utilizing SNNs trained with dynamic sparse training. We show that SNNs trained with DST can achieve stable feature selection results regardless of the training algorithm and considered metric, even outperforming the dense neural network feature selection performance in most cases considered, while significantly reducing memory and FLOPs. 

    \item We introduce a novel feature importance metric based on neuron attribution, highlighting its advantages over existing feature importance methods in capturing the feature relevance.

\end{itemize}

\looseness=-1
This paper aims to deepen the understanding of feature selection using sparse neural networks and provide a robust framework for evaluating various feature importance metrics and training algorithms in feature selection with sparse neural networks and DST framework. It should be noted that in this work, we focus on global feature selection rather than local feature selection (e.g. instance-based feature selection, LIME, local feature importance).

\section{Backgound \& Related Work}\label{sec:realted_work}

\subsection{Feature Selection}
Methods to perform feature selection are divided into three main categories: Filter, wrapper, and Embedded methods. \textbf{Filter Methods} \cite{guyon2003introduction, he2006laplacian, chandrashekar2014survey} exploit variable scoring techniques to rank the features. They are model-agnostic and do not rely on the learning algorithm. As the ranking is done before classification, filter methods are prone to selecting irrelevant features. \textbf{Wrapper Methods} \cite{zhang2019feature, kohavi1997wrappers, liu1996probabilistic} aim to find a subset of the feature that achieves the highest classification performance. To tackle the NP-hard problem of evaluating numerous subsets, they employ search algorithms to find the best subset. However, due to the multiple rounds of training, these methods are costly in terms of computation. \textbf{Embedded Methods} integrate feature selection into the learning process thus being able to select relevant features while being costly-efficient. Various techniques are employed to perform embedded feature selection including, mutual information \cite{battiti1994using, peng2005feature}, the SVM classifier \cite{guyon2002gene}, and neural networks \cite{setiono1997neural}. 

Neural network-based feature selection has gained significant attention in recent years, both in supervised \cite{lu2018deeppink, lemhadri2021lassonet, yamada2020feature, wojtas2020feature, jia2023effective, espinosa2023embedded, peng2023copula} and unsupervised \cite{balin2019concrete, han2018autoencoder,chandra2015exploring, doquet2019agnostic} settings. These methods leverage the advantages of neural networks in capturing non-linear dependencies and performing well on large datasets. However, many existing neural network-based feature selection methods suffer from over-parameterization, resulting in high computational costs, especially for high-dimensional datasets. To address these issues, a new category of methods exploits sparse neural networks to perform efficient feature selection \cite{atashgahi2022quick, sokar2022pay, atashgahi2022supervised}. Detailed discussion on SNNs for feature selection can be found in Section \ref{sssec:dst_fs_background}.

\subsubsection{Problem Formulation} 
Depending on label availability, feature selection can be carried out in a supervised or unsupervised manner. This study focuses on addressing the supervised feature selection challenge. Given a dataset $\sX$ comprising $m$ samples denoted as $(\vx^{(i)}, y^{(i)})$, where $\vx^{(i)} \in \mathbb{R}^d$ represents the $i$-th sample in the data matrix $\mX \in \mathbb{R}^{m \times d}$ (with $d$ being the dataset's dimensionality or the number of features) and $y^{(i)}$ is the corresponding label for supervised learning, the objective is to choose a subset of the most discriminative and informative features from $\mX$, denoted as $\mathbb{F}_s \subset \mathbb{F}$. Here, $|\mathbb{F}_s|=K$, with $\mathbb{F}$ being the original feature set, and $K$ is a hyperparameter indicating the number of selected features. In supervised feature selection, we seek to optimize:

\begin{equation}\label{eq:supervised_FS}
     \mathbb{F}^{*}_s = \argmin_{ \mathbb{F}_s \subset  \mathbb{F}, | \mathbb{F}_s|=K} \sum\limits_{i=0}^{m-1} J (f(\vx^{(i)}_{\mathbb{F}_s} ; \vtheta), y^{(i)}),
\end{equation}
where $\mathbb{F}^{*}_s$ is the final selected feature set, $J$ is a desired loss function, and $f(\vx^{(i)}_{\mathbb{F}_s} ; \vtheta)$ is a classification function parameterized by $\vtheta$ estimating the target for the $i$-th sample using a feature subset $\vx^{(i)}_{\mathbb{F}_s}$. However, directly optimizing Equation \ref{eq:supervised_FS} is an NP-hard problem \cite{lemhadri2021lassonet}. Therefore, by estimating the importance of the features using the neuron attribution method, we try to find the closest set of features to the optimal feature set found by Equation \ref{eq:supervised_FS}.

\begin{figure}[!t]
  \centering

  \includegraphics[width=0.4\textwidth]{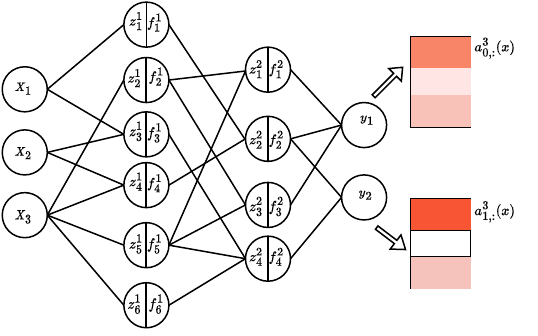} 
  \vspace{2mm}
  \caption{Neuron Attribution visualization in a sparse neural network. The contribution of each input feature for any output neuron is measured by neuron attribution methods. Darker colors show a higher contribution of the corresponding input neuron to the output neuron.}
  \label{fig:neuron_attribution}
  \vspace{3mm}
\end{figure}

\subsection{Neuron Attribution}\label{ssec:back_neuron_attribtuion}
In this work, we propose a new feature selection metric from neural networks that is based on neuron attribution. Attribution methods aim to explain the predictions of a neural network. These methods have also been used to explain the predictions of a model\cite{ancona2018towards}. Many works have regularized network attribution to achieve the desired output and improve the learning of neural networks \cite{rifai2011contractive, gulrajani2017improved, moosavi2019robustness, jeffares2023tangos}. Given a neural network, an attribution method aims to determine the contribution/relevancy of each input feature to the output of $i-th$ output neuron ($f^{L-1}_i$) \cite{ancona2018towards}. Usually, the gradient information is used to measure the neuron attribution. Neuron attribution for the $i^{th}$ neuron in the output layer w.r.t. the feature $x_j$ can be simply defined as:
\begin{equation}\label{eq:neuron_attribution}
    a^{L-1}_{ij}(x) =\frac{\partial f^{L-1}_i(x)}{\partial x_j}, 
\end{equation}

where $L$ is the number of network layers and $L-1$ is the index of the output layer. The higher the absolute value of $a^{L-1}_{ij}(x)$ is, the more sensitive output neuron $i$ is to the changes in the input feature $x_j$ for observation $x$. Other gradient attribution methods can be used to measure the neuron attribution in Equation \ref{eq:neuron_attribution} \cite{ancona2018towards}. An example of neuron attribution attribution is visualized in Figure \ref{fig:neuron_attribution}.

\subsection{Sparse Neural Networks}
Sparse neural networks are an approach to address the overparameterization of deep neural networks. By introducing sparsity in the connectivity and/or the units of a dense neural network, they aim to match the performance of dense neural networks while reducing computational and memory costs \cite{hoefler2021sparsity, mocanu2021sparse}. Due to the reduction of learned noise, they can even improve the generalization of the network \cite{hoefler2021sparsity}. A sparse neural network is represented by $f(\vx, \vtheta_s)$ that has a sparsity level of $P$, where $P$ is calculated as $1 - \frac{\norm{\vtheta_s}_0}{\norm{\vtheta}_0}$. Here, $\vtheta_s$ denotes a subset of parameters of the dense network parameterized by $\vtheta$, and $\norm{\vtheta_s}_0$ and $\norm{\vtheta}_0$ refer to the number of parameters of the sparse and dense network respectively.

\begin{table}[!b]
\centering
\caption{Comparison with related works that exploit sparsity for feature selection.}
\vspace{2mm}
\label{tab:related_work}
\scalebox{1}{
    \begin{tabular}{@{\hskip 0.0in}c@{\hskip 0.07in}c@{\hskip 0.07in}c@{\hskip 0.07in}c@{\hskip 0.07in}cccc}
    \toprule
    \textbf{Method} & \textbf{Sparsity}  &\textbf{Scalability} &\makecell{\textbf{Neuron Attribution} \\ \textbf{Importance}}\\
    \midrule
    Lasso \cite{tibshirani1996regression}& Regularization  &\checkmark&\xmark\\\
    STG \cite{yamada2020feature} & Regularization &\xmark&\xmark\\
    LassoNet \cite{lemhadri2021lassonet} & Regularization &\xmark&\xmark\\
    QuickSelection \cite{atashgahi2022quick} & DST  &\checkmark&\xmark\\
    SET-Attr (ours) & DST  &\checkmark&\checkmark\\

    \bottomrule
    \end{tabular}
}

\end{table}

\subsubsection{Dynamic Sparse Training (DST)}\label{sssec:dst}
DST comprises a range of techniques to train sparse neural networks from scratch. The goal of DST methods is to optimize the sparse connectivity of the network during training, without resorting to dense network matrices at any point \cite{mocanu2021sparse, mocanu2018scalable, yuan2021mest}. To achieve this, DST methods begin with initializing a random sparse neural network. During training, DST methods periodically update the sparse connectivity of the network by removing a fraction $\zeta$ of the parameters $\vtheta_s$ and adding the same number of parameters to the network to keep the sparsity level fixed. In the literature, usually, weight magnitude has been used as a criterion for dropping the connections. However, there exist various approaches for weight regrowth including, random \cite{mocanu2018scalable}, gradient \cite{evci2020rigging, jayakumar2020top}, and neuron similarity \cite{atashgahi2022brain}.

\begin{figure}[!t]
  \centering

  \includegraphics[width=0.25\textwidth]{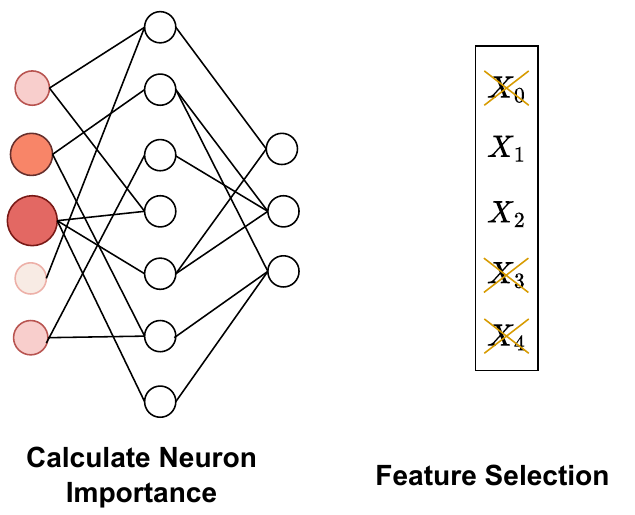} 
  \vspace{2mm}
  \caption{Feature selection using sparse neural network. The importance of each input neuron can be measured using the network's characteristics. Darker colors and larger neurons show a higher importance of the corresponding input neuron.}
  \label{fig:FS}
  \vspace{3mm}
\end{figure}

\subsubsection{DST for Feature Selection}\label{sssec:dst_fs_background} 
QuickSelection was the first work to show that sparse neural networks trained with DST can perform feature selection. It proposed the \emph{neuron strength} \cite{atashgahi2022quick} in a sparse denoising autoencoder to determine the importance of each input neuron in the input layer of a sparse neural network (and the corresponding input feature in the original dataset) in the unsupervised learning settings. Neuron strength is computed as:

\begin{equation}\label{eq:imp_input_neuron_strength}
S(X_j)= \sum_{i=0}^{n^{h^1}-1}{|W^1_{ji}|},
\end{equation}

where $n^{h^1}$ is the number of hidden neurons in the first hidden layer, and $W^l$ is the weight matrix of the $l$th layer. 

Later on, \cite{sokar2022pay}, proposed to use a combination of neuron strength and gradient of loss to the output neurons in an autoencoder to measure the importance of neurons. \cite{atashgahi2022supervised} proposed to perform input neuron updating in a sparse MLP trained with DST to gradually reduce the number of input neurons and then use neuron strength to rank the features.

In this paper, we study the efficacy of SNNs for feature selection and propose a new importance metric based on neuron attribution to measure the importance of input features in neural networks. Table \ref{tab:related_work}, compares our proposed method with the closest related work that exploit sparsity to perform feature selection. Figure \ref{fig:FS} shows a toy example of feature selection using sparse neural networks. 

\section{Methodology}\label{sec:methodology}
The contributions of this research are twofold. Firstly, we provide an extensive analysis of the performance of SNNs trained with DST for feature selection. Secondly, we propose a new metric to derive the importance of features in an SNN trained with DST. We elaborated on the experimental analysis settings and findings in Section \ref{sec:results}. In this section, we describe our proposed approach to measure neuron/feature importance based on neuron attribution. 

\looseness=-1
\textbf{Motivation.} As elucidated in Section \ref{ssec:back_neuron_attribtuion}, neuron attribution serves as a metric to estimate the relevance of an input feature to a specific output neuron, given an input sample. Consequently, the neuron attribution vector for a hidden neuron provides insights into which set of features is more relevant to derive the output. Looking at the contribution vectors of all output neurons, we can select the most relevant input features for each output feature. Thus, neuron attribution enables us to rank input features based on their relevance to the output features. Features with the highest contributions in the output neurons offer a robust estimate of the output, implicitly guiding us toward identifying the optimal feature set, as represented in Equation \ref{eq:supervised_FS}. Building on this premise, we introduce the neuron attribution feature selection metric in the context of neural networks.

\begin{figure}[!t]
  \centering

  \includegraphics[width=0.35\textwidth]{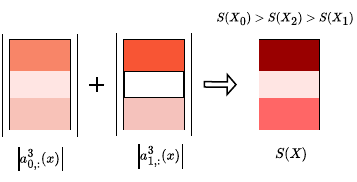} 
  \vspace{2mm}
  \caption{A toy example of neuron attribution-based importance calculation. Darker colors indicate higher contributions of the corresponding input neuron.}
  \label{fig:neuron_importance_s}
  \vspace{3mm}
\end{figure}
\subsection{Input Neuron Importance}\label{ssec:neuron_importance} 
\looseness=-1
The importance of the input neurons is computed based on the neuron attribution of the output neurons. We propose to compute the importance score of each input neuron as the following: 

\begin{equation}\label{eq:imp_input_neuron_attribution}
S_{itr}(X_j)=\sum_{i=0}^{C-1}[\frac{1}{m}\sum_{k=0}^{m-1}|a^{L-1}_{ij}(x_k)|],
\end{equation}

\looseness=-1
where $S_{itr}(X_j)$ is the importance of the $j$th feature at iteration $itr$, $C$ is the number of output neurons (classes), $m$ is the number of samples, and $a^{L-1}_{ij}(x_k)$ is the neuron attribution of the $j$th input neuron for the $i$th output neuron and sample $x_k$. To compute neuron importance in Equation \ref{eq:imp_input_neuron_attribution} for each input neuron, we sum the neuron attribution for all output neurons (averaged over all samples); this choice is driven by empirical results. In other words, a feature is important if it is highly relevant for most output features and input samples. For computing the overall importance of the neurons during training, we sum $S_{itr}(X_j)$ over all the training iterations. While attribution methods have been studied only in dense neural networks, in this paper, we study these methods for sparse neural networks. A toy example of this process is visualized in Figure \ref{fig:neuron_importance_s}.

 \textbf{Feature Selection.} After deriving the input neurons importance based in Equation \ref{eq:imp_input_neuron_attribution}, we select the features corresponding to the neurons with the highest neuron attribution-based importance as the final feature set. 
\section{Results}\label{sec:results}
In this section, we first describe the experimental settings (\ref{ssec:settings}). Then, we present the results of the feature selection comparison among sparse and dense models in Section \ref{ssec:fs_100_comparison}, and among standard feature selection baselines using sparsity in Section \ref{ssec:sparsity_comparison}. Finally, we visualize the neuron importance in Section \ref{ssec:neuron_imp_visualization}. Additionally, we perform an analysis on a synthetic dataset in Appendix \ref{app:synthetic}.

\begin{table}[!t]
\centering
\caption{Datasets characteristics.}
\vspace{2mm}
\label{tab:datasets}
\scalebox{0.95}{
    \begin{tabular}{@{\hskip 0in}c@{\hskip 0.07in}c@{\hskip 0.07in}c@{\hskip 0.07in}c@{\hskip 0.07in}c@{\hskip 0in}}
    \toprule
     \textbf{Type}  &\textbf{Dataset} &\textbf{\#Samples}  &\textbf{\#Features}      &\textbf{\#Classes} \\
    \midrule
    
        \multirow{3}{*}{Image (Hand-written)} &MNIST  &70000    &784  &10\\
        &USPS   &9298	&256	  & 10\\
        &Gisette    &7000	&5000	&2\\\midrule
        \multirow{3}{*}{Image} &COIL20	&1440	&1024	&20\\
        &ORL	&400	&1024	&40\\
        &Yale	&165	&1024	&15\\\midrule
        \multirow{3}{*}{Text} &BASEHOCK &1993 & 4862 &2\\
        & PCMAC  &1943  &3289 &2\\
        & RELATHE	&1427	&4322	&2\\\midrule
        \multirow{5}{*}{Biological}  &Prostate-GE & 102 &5966  &2\\
        &SMK-CAN-187   & 187 &  19993   & 2\\  
        & CLL-SUB-111  &111 &11340   &3\\
        &lymphoma	&96	&4026	&9\\
        &Carcinom  &174  &9182    &11\\\midrule
        Time Series  &HAR  & 10299  &561   &6\\\midrule
        Mass Spectrometry &Arcene &200 &10000   &2\\\midrule
        Speech &Isolet	&1560	&617	&26\\\midrule
        Artificial (Noisy)&Madelon	&2600	&500	&2\\

    \bottomrule
    \end{tabular}
}
\end{table}

\begin{table*}[h]
\caption{Feature selection results in terms of test classification accuracy [\%] of an SVM classifier on the selected subset of (K=100). The values in the parenthesis show the sparsity level.}
\vspace{2mm}
\begin{subtable}[h]{0.3\textwidth}
        \centering
        \scalebox{1}{
        \begin{tabular}{c@{\hskip 0.07in}c@{\hskip 0.07in}c@{\hskip 0.07in}c}
        \toprule

            & \textbf{Dense-Attr} & \textbf{SET-Attr}\\\midrule

MNIST & 96.23 $\pm$ 0.08 (0.00) & \textbf{96.24 $\pm$ 0.13} (0.25)\\
USPS & \textbf{96.83 $\pm$ 0.13} (0.00) & 96.77 $\pm$ 0.13 (0.50)\\
Gisette & 97.03 $\pm$ 0.26 (0.00) & \textbf{97.10 $\pm$ 0.23} (0.50)\\
\midrule
Coil20 & \textbf{98.82 $\pm$ 1.17} (0.00) & 98.02 $\pm$ 1.85 (0.50)\\
ORL & 89.38 $\pm$ 3.80 (0.00) & \textbf{89.50 $\pm$ 2.45} (0.25)\\
Yale & 63.94 $\pm$ 7.48 (0.00) & \textbf{65.76 $\pm$ 7.05} (0.25)\\
\midrule
BASEHOCK & 92.83 $\pm$ 1.51 (0.00) & \textbf{94.34 $\pm$ 0.66} (0.95)\\
PCMAC & 84.96 $\pm$ 1.40 (0.00) & \textbf{87.89 $\pm$ 1.26} (0.95)\\
RELATHE & 80.24 $\pm$ 2.38 (0.00) & \textbf{82.10 $\pm$ 1.02} (0.95)\\
\midrule
Prostate\_GE & \textbf{89.05 $\pm$ 2.18} (0.00) & 88.57 $\pm$ 2.33 (0.50)\\
SMK & \textbf{81.05 $\pm$ 3.07} (0.00) & 80.53 $\pm$ 2.41 (0.50)\\
CLL & \textbf{83.48 $\pm$ 5.43} (0.00) & 78.26 $\pm$ 8.02 (0.25)\\
Carcinom & \textbf{84.57 $\pm$ 4.64} (0.00) & 83.43 $\pm$ 5.83 (0.50)\\
Lymphoma & 64.50 $\pm$ 5.68 (0.00) & \textbf{66.00 $\pm$ 7.35} (0.50)\\
\midrule
Arcene & 65.25 $\pm$ 7.62 (0.00) & \textbf{67.75 $\pm$ 7.11} (0.25)\\
\midrule
HAR & 94.78 $\pm$ 0.37 (0.00) & \textbf{94.90 $\pm$ 0.39} (0.25)\\
\midrule
Isolet & 94.33 $\pm$ 1.31 (0.00) & \textbf{95.53 $\pm$ 0.42} (0.95)\\
\midrule
Madelon & \textbf{81.52 $\pm$ 2.55} (0.00) & 76.63 $\pm$ 2.36 (0.95)\\
\midrule

       \end{tabular}}
       \caption{Dense-Attr vs. SET-Attr.}
       \label{tab:denseVSsparse}
    \end{subtable}
    \hfill\hfill\hfill\hfill\hfill\hfill\hfill\hfill\hfill\hfill\hfill\hfill\hfill
    \begin{subtable}[h]{0.3\textwidth}
        \centering
        \scalebox{1}{
        \begin{tabular}{c@{\hskip 0.0in}c@{\hskip 0.07in}c@{\hskip 0.07in}c}
        \toprule
            & \textbf{RigL-Attr} & \textbf{SET-Attr} \\\midrule

 & 96.22 $\pm$ 0.10 (0.50) & \textbf{96.24 $\pm$ 0.13} (0.25)\\
 & \textbf{96.87 $\pm$ 0.17} (0.80) & 96.77 $\pm$ 0.13 (0.50)\\
 & 97.07 $\pm$ 0.27 (0.50) & \textbf{97.10 $\pm$ 0.23} (0.50)\\
\midrule
 & \textbf{98.06 $\pm$ 1.87} (0.80) & 98.02 $\pm$ 1.85 (0.50)\\
 & \textbf{90.88 $\pm$ 3.26} (0.25) & 89.50 $\pm$ 2.45 (0.25)\\
 & \textbf{73.94 $\pm$ 5.94} (0.50) & 65.76 $\pm$ 7.05 (0.25)\\
\midrule
 & 86.69 $\pm$ 2.20 (0.95) & \textbf{94.34 $\pm$ 0.66} (0.95)\\
 & 81.70 $\pm$ 2.29 (0.95) & \textbf{87.89 $\pm$ 1.26} (0.95)\\
 & 74.69 $\pm$ 2.96 (0.98) & \textbf{82.10 $\pm$ 1.02} (0.95)\\
\midrule
 & 88.09 $\pm$ 3.19 (0.80) & \textbf{88.57 $\pm$ 2.33} (0.50)\\
 & 80.26 $\pm$ 3.95 (0.50) & \textbf{80.53 $\pm$ 2.41} (0.50)\\
 & \textbf{78.70 $\pm$ 7.64} (0.25) & 78.26 $\pm$ 8.02 (0.25)\\
 & 80.00 $\pm$ 6.26 (0.50) & \textbf{83.43 $\pm$ 5.83} (0.50)\\
 & 65.50 $\pm$ 7.57 (0.50) & \textbf{66.00 $\pm$ 7.35} (0.50)\\
\midrule
 & 57.00 $\pm$ 6.40 (0.80) & \textbf{67.75 $\pm$ 7.11} (0.25)\\
\midrule
 & 94.88 $\pm$ 0.40 (0.50) & \textbf{94.90 $\pm$ 0.39} (0.25)\\
\midrule
 & 94.72 $\pm$ 0.58 (0.90) & \textbf{95.53 $\pm$ 0.42} (0.95)\\
\midrule
 & 75.75 $\pm$ 3.50 (0.98) & \textbf{76.63 $\pm$ 2.36} (0.95)\\
\midrule

       \end{tabular}}
       \caption{RigL-Attr vs. SET-Attr.}
       \label{tab:RigLvsSET}
    \end{subtable}
    \hfill
    \begin{subtable}[h]{0.3\textwidth}
        \centering
        \scalebox{1}{
        \begin{tabular}{c@{\hskip 0.0in}c@{\hskip 0.07in}c@{\hskip 0.07in}c}
        \toprule

            & \textbf{SET-QS} & \textbf{SET-Attr}\\\midrule

 & 96.02 $\pm$ 0.10 (0.25) & \textbf{96.24 $\pm$ 0.13} (0.25)\\
 & 96.64 $\pm$ 0.13 (0.50) & \textbf{96.77 $\pm$ 0.13} (0.50)\\
 & 96.65 $\pm$ 0.45 (0.50) & \textbf{97.10 $\pm$ 0.23} (0.50)\\
\midrule
 & \textbf{98.72 $\pm$ 0.31} (0.50) & 98.02 $\pm$ 1.85 (0.50)\\
 & \textbf{92.25 $\pm$ 2.15} (0.25) & 89.50 $\pm$ 2.45 (0.25)\\
 & \textbf{74.55 $\pm$ 3.64} (0.25) & 65.76 $\pm$ 7.05 (0.25)\\
\midrule
 & 93.78 $\pm$ 1.01 (0.95) & \textbf{94.34 $\pm$ 0.66} (0.95)\\
 & 86.76 $\pm$ 1.41 (0.95) & \textbf{87.89 $\pm$ 1.26} (0.95)\\
 & 80.77 $\pm$ 1.35 (0.95) & \textbf{82.10 $\pm$ 1.02} (0.95)\\
\midrule
 & 87.62 $\pm$ 4.86 (0.50) & \textbf{88.57 $\pm$ 2.33} (0.50)\\
 & 79.47 $\pm$ 1.97 (0.50) & \textbf{80.53 $\pm$ 2.41} (0.50)\\
 & 76.09 $\pm$ 8.53 (0.25) & \textbf{78.26 $\pm$ 8.02} (0.25)\\
 & \textbf{89.14 $\pm$ 2.49} (0.50) & 83.43 $\pm$ 5.83 (0.50)\\
 & \textbf{70.50 $\pm$ 6.10} (0.50) & 66.00 $\pm$ 7.35 (0.50)\\
\midrule
 & 66.25 $\pm$ 8.31 (0.25) & \textbf{67.75 $\pm$ 7.11} (0.25)\\
\midrule
 & 94.59 $\pm$ 0.41 (0.25) & \textbf{94.90 $\pm$ 0.39} (0.25)\\
\midrule
 & 95.22 $\pm$ 0.35 (0.95) & \textbf{95.53 $\pm$ 0.42} (0.95)\\
\midrule
 & \textbf{78.40 $\pm$ 1.71} (0.95) & 76.63 $\pm$ 2.36 (0.95)\\
\midrule

       \end{tabular}}
       \caption{SET-QS vs. SET-Attr.}
       \label{tab:AttrVSQS}
    \end{subtable}
    \hfill

     \label{tab:results_fs}
\end{table*}

\subsection{Experimental Setup}\label{ssec:settings}
In the following, we describe datasets, baselines, and the evaluation approach. More experimental details, including the hyperparameter settings and model architecture, can be found in Appendix \ref{app:settings}.
\subsubsection{Datasets}
The datasets, outlined in Table \ref{tab:datasets}, include a diverse collection of $18$ datasets, varying in size and type. This selection allows for a comprehensive analysis of each method across different domains. More than half of these datasets are high-dimensional making them challenging benchmarks for the models.

\subsubsection{Baselines}\label{sssec:baselines}
\textbf{DST for Feature Selection.} The main focus of this paper is to analyze how sparse neural networks trained with DST perform in feature selection compared to dense networks. Therefore, we consider dense and sparse baselines. Secondly, for the sparse models, we want to study how the DST algorithm for training the sparse neural network affects the feature selection performance. We consider two standard DST approaches in the literature, SET \cite{mocanu2018scalable} and RigL \cite{evci2020rigging}, that are frequently used in evaluating the DST framework. We describe SET and RigL in Appendix \ref{s-app:settings:baselines}. Finally, we want to assess the efficacy of the neuron importance metric. We consider the neuron strength metric from QuickSelection \cite{atashgahi2022quick}, which is also used as a part of neuron importance estimation in \cite{sokar2022pay} or directly used in \cite{atashgahi2022supervised} to rank the features in a supervised setting; we call this metric as \enquote{QS} in the experiments. We compare the neuron strength metric with our proposed neuron attribution metric, called \enquote{Attr}. These metrics combined with the three considered models, dense network (\emph{Dense}), SNN trained with SET (\emph{SET}), and SNN trained with RigL (\emph{RigL}), result in $6$ baselines for the experiments: \emph{Dense-QS}, \emph{Dense-Attr}, \emph{SET-QS}, \emph{SET-Attr}, \emph{RigL-QS}, \emph{RigL-Attr}. For QuickSelection, as we also do for Attr, to compute the input neuron/feature importance, we sum the importance during all training epochs. As shown in Appendix \ref{app:importance_metric}, we observed that looking at the history of importance during all training epochs improves the results for these models.

\textbf{Sparsity for Feature Selection.} We consider three feature selection methods that exploit sparsity to perform feature selection including STG \cite{yamada2020feature}, LassoNet \cite{lemhadri2021lassonet}, and Lasso \cite{tibshirani1996regression}. 

\subsubsection{Evaluation}
For evaluating each method, we first perform feature selection to derive $K$ most important features ($K$ is a hyperparameter set by the user). Then, we train an SVM classifier on the subset of K features of the original data and report the test accuracy on a hold-out test set.

\subsection{Feature Selection Comparison}\label{ssec:fs_100_comparison}

\textbf{Settings.} In this experiment, we compare the two feature ranking criteria, neuron strength from QuickSelection (QS), and neuron attribution (Attr). We consider dense and sparse MLPs to perform feature selection on. The sparse models are trained with DST; we consider SET and RigL as the training algorithms. The comparison results are summarized in Tables \ref{tab:results_fs}. For ease of comparison, we considered pairwise comparison. However, comparisons of all baselines referred to in Section \ref{sssec:baselines}, are brought in Appendix \ref{app:more_comparison}. Additionally, we analyze the performance of each method across various $K$ values in $\{25, 50, 75, 100, 200\}$ in terms of average ranking in Appendix \ref{app:K_values}.

\textbf{Dense vs Sparse.}
Comparing dense and sparse models, we consider Dense-Attr and SET-Attr. In Table \ref{tab:denseVSsparse}, we can observe that while on high-dimensional biological datasets and the noisy dataset (Madelon) Dense-Attr performs better, on the rest of the datasets (11 out of 18 datasets) SET-Attr excels the dense model. Additionally, SET-Attr exploits significantly fewer parameters (memory) and FLOPs (computational costs). As shown in Table \ref{tab:results_fs}, on average over all datasets, SET achieves $54.2\%$ sparsity and RigL achieves $66.4\%$ sparsity. In Table \ref{tab:results_flops}, we present the FLOPs (Floating-Point Operations) count for SET-Attr and Dense-Attr; SET significantly reduces the number of FLOPs, and as a result, computational costs, in all cases compared to the dense model.

\looseness=-1
\textbf{DST Algorithm.}
Table \ref{tab:RigLvsSET} compares the results of RigL-Attr and SET-Attr. While on the Image datasets (Coil20, ORL, and Yale), RigL outperforms SET in the quality of the selected features on the rest of the datasets SET outperforms RigL or has comparable performance. On average, RigL achieves $66.4\%$ sparsity, while SET achieves $54.2\%$ sparsity.

\looseness=-1
\textbf{Neuron Attribution vs Neuron Strength.}
In Table \ref{tab:AttrVSQS}, we compare the two neuron strength (QS) and neuron attribution (Attr) metrics, when the network is trained using the SET algorithm. Across the hand-written datasets (MNIST, USPS, Gisette), text datasets (BASEHOCK, PCMAC, RELATHE), and HAR, SET-Attr consistently yield better results than SET-QS on each of the models. This pattern is similar when the network is trained as Dense or with RigL (please see Table \ref{tab:results_fs_more} in Appendix \ref{app:more_comparison}). In image datasets (Coil20, ORL, Yale) and Madelon, SET-QS outperforms SET-Attr. The Biological datasets (Prostate\_GE, SMK, CLL, Carcinom, Lymphoma) showcase varying degrees of performance for different methods. SET-Attr performs better in some cases, such as Prostate\_GE, SMK, and CLL, while in others, such as Carcinom and Lymphoma, SET-QS shows superiority. This indicates the complexity and diversity of biological data, suggesting that the choice of feature selection method might depend heavily on the specific dataset characteristics.

\textbf{Conclusions.} The results show that neuron attribution generally outperforms neuron strength, sparse models outperform dense ones, and SET is generally a better choice when training the sparse neural networks for feature selection than RigL. However, selecting an appropriate feature selection metric and training algorithm based on dataset characteristics and domain requirements is of great importance.

\subsection{Neuron Importance Visualization}\label{ssec:neuron_imp_visualization}
\begin{figure}[!b]
    \centering
    \begin{subfigure}[t]{0.15 \textwidth}
        \centering
        \includegraphics[width=\textwidth]{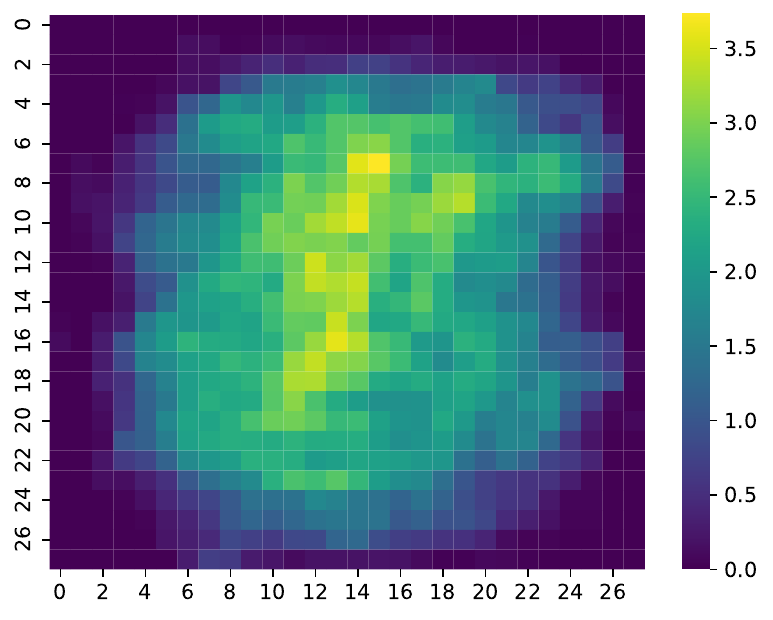}
        \caption{Dense-QS}\label{fig:input_dense_QS}
    \end{subfigure}
    \begin{subfigure}[t]{0.15 \textwidth}
        \centering
        \includegraphics[width=\textwidth]{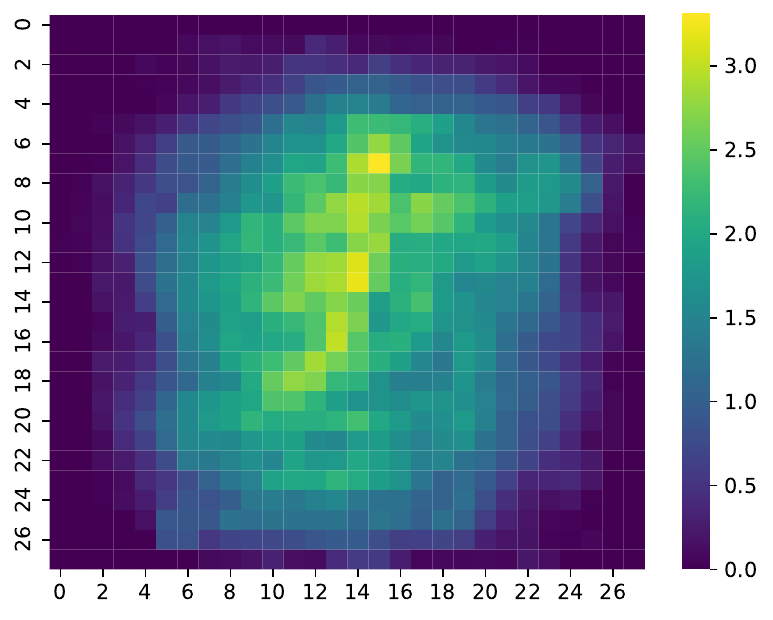}
        \caption{SET-QS}\label{fig:input_set_QS}
    \end{subfigure}
    \begin{subfigure}[t]{0.15 \textwidth}
        \centering
        \includegraphics[width=\textwidth]{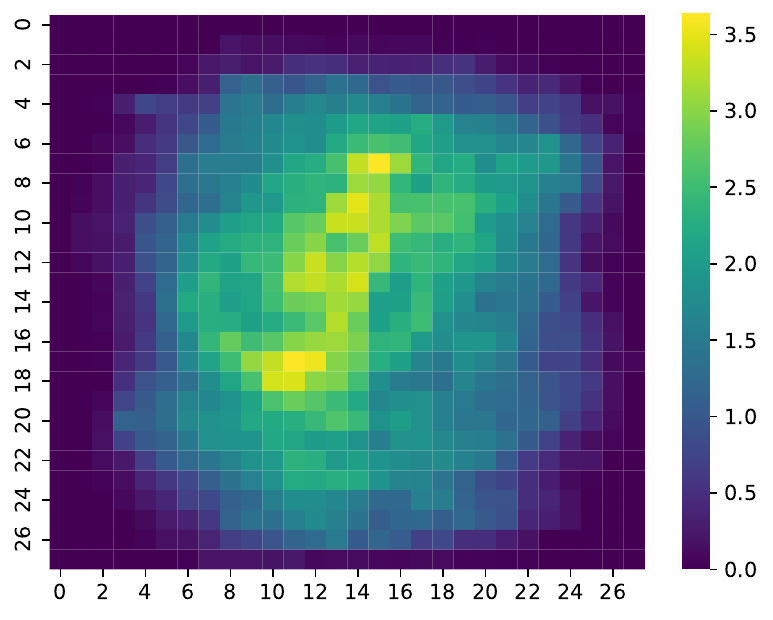}
        \caption{RigL-QS}\label{fig:input_rigl_QS}
    \end{subfigure}
    \begin{subfigure}[t]{0.15 \textwidth}
        \centering
        \includegraphics[width=\textwidth]{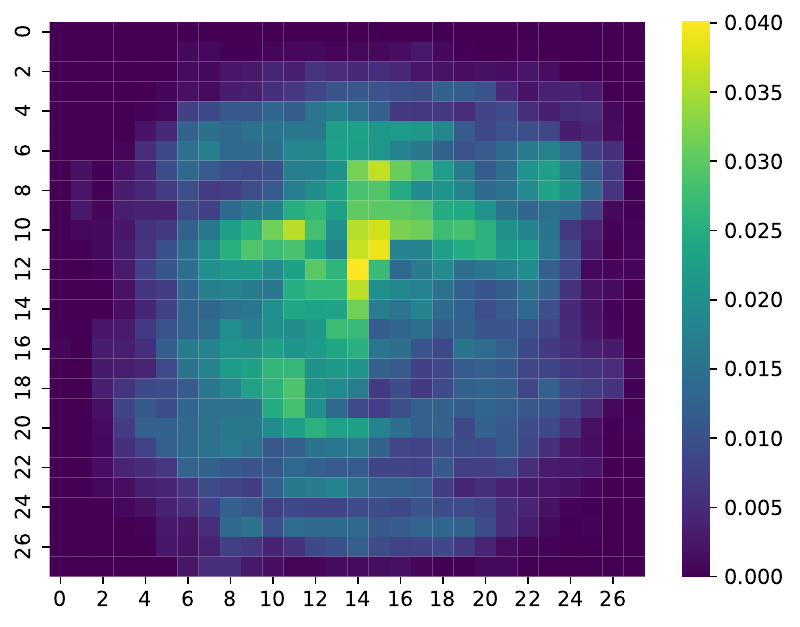}
        \caption{Dense-Attr}\label{fig:input_dense_Attr}
    \end{subfigure}
    \begin{subfigure}[t]{0.15 \textwidth}
        \centering
        \includegraphics[width=\textwidth]{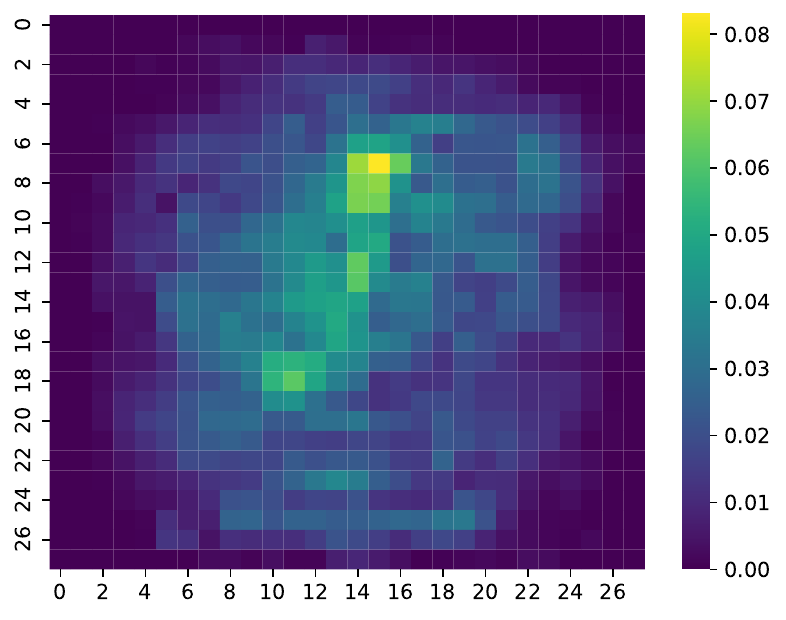}
        \caption{SET-Attr}\label{fig:input_set_Attr}
    \end{subfigure}
    \begin{subfigure}[t]{0.15 \textwidth}
        \centering
        \includegraphics[width=\textwidth]{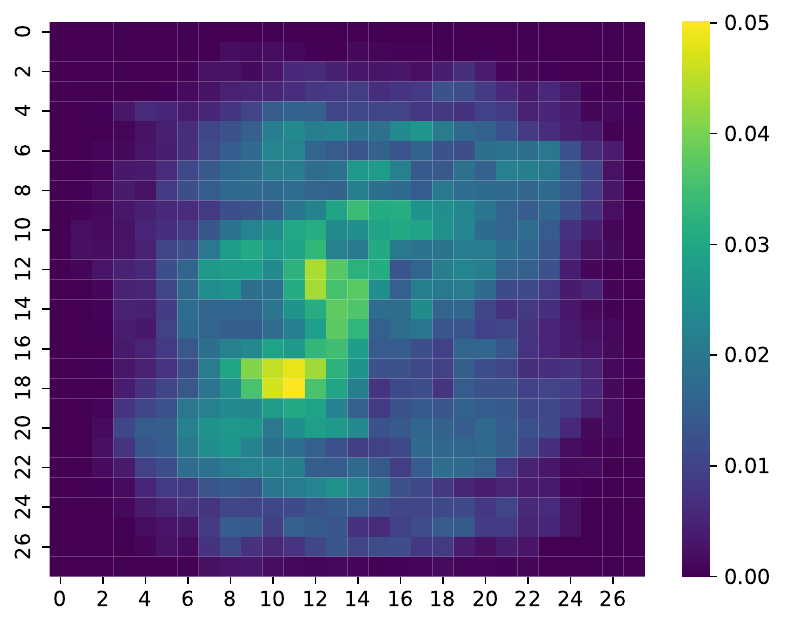}
        \caption{RigL-Attr}\label{fig:input_rigl_Attr}
    \end{subfigure}
    \vspace{2mm}
    \caption{Neuron importance visualization on the MNIST dataset as 2d-heat maps. The lighter area in the center of the Figures shows more important features which is in-line with the pattern of digits in the MNIST dataset.}\label{fig:neuron_importance}
\end{figure}

\looseness=-1
\textbf{Settings.} In this section, we visualize the neuron importance on the MNIST dataset for each method in Table \ref{tab:results_fs}. We plot the neuron importance of input features as a 2D heatmap in Figure \ref{fig:neuron_importance}.

\textbf{Results.} As we can observe in the first row of Figure \ref{fig:neuron_importance}, the neuron strength metric in QuickSelection is mostly similar for all three models, showing that they all can detect the most important features which mostly appear in the center of the image as we see in the MNIST dataset. The pattern formed by the neuron attribution metric in the second row of Figure \ref{fig:neuron_importance}, is close to that of neuron strength, where important features appear in the center of the image. However, the main difference is that neuron attribution reaches a more sparse feature importance pattern. This shows that neuron attribution focuses mostly on the most important features.

 \begin{figure*}[!t]
  \centering
  \includegraphics[width=\textwidth]{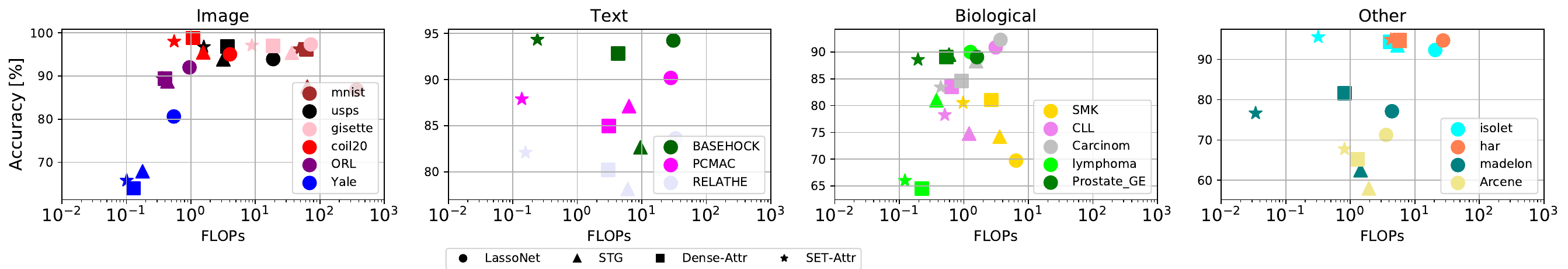} 
  \vspace{1mm}
  \caption{Accuracy vs. FLOPs comparison among various neural network-based feature selection methods inducing sparsity in the network. The FLOPs values are divided by $10^{12}$}
  \label{fig:acc_flops}
\end{figure*}

\subsection{Comparsion with Baselines}\label{ssec:sparsity_comparison}

\textbf{Settings.}
We compare SET-Attr (achieving the highest ranking among the considered methods when $K=100$) with three popular feature selection methods in the literature that exploit sparsity to perform feature selection: Lasso, STG \cite{yamada2020feature}, and LassoNet \cite{lemhadri2021lassonet}. We compare the feature selection performance of these methods when selecting $K=100$ features.

\textbf{Feature Selection Performance.}
The feature selection results are summarized in Table \ref{tab:results_fs_sparse}. SET-Attr and LassoNet are the best performers among the considered methods, each being 7 and 10 times the best performers, respectively. STG and Lasso consistently fall behind both SET-Attr and LassoNet in most datasets. By looking at the theoretical estimation of FLOPs count in Table \ref{tab:results_flops}, we can observe that SET-Attr has the lowest number of FLOPs in all datasets, having on average $88.2\%$ less flops than LassoNet. This indicates the efficiency of SET-Attr while achieving a competitive feature selection performance to the SOTA feature selection algorithms.

\begin{table}[!b]

    \centering
    \caption{Feature selection results in terms of classification accuracy of an SVM classifier on the selected subset of features (K=100). \textbf{Bold} entries denote the best performers.}
    \vspace{2mm}
    \label{tab:results_fs_sparse}
    \centering
            \scalebox{1}{
             \begin{tabular}{c@{\hskip 0.07in}c@{\hskip 0.07in}c@{\hskip 0.07in}c@{\hskip 0.07in}c@{\hskip 0.07in}c@{\hskip 0.07in}c@{\hskip 0.07in}c@{\hskip 0.07in}c@{\hskip 0.07in}c@{\hskip 0.07in}c@{\hskip 0.07in}c@{\hskip 0.07in}c@{\hskip 0.07in}}
                \toprule

& \textbf{Dataset}  & \textbf{Lasso} & \textbf{LassoNet}  & \textbf{STG}  & \textbf{SET-Attr}\\\midrule

 & MNIST & 78.27 $\pm$ 0.00 & 86.83 $\pm$ 0.11 & 87.54 $\pm$ 0.55 & \textbf{96.24 $\pm$ 0.13}\\
 & USPS & 93.55 $\pm$ 0.00 & 93.92 $\pm$ 0.16 & 93.88 $\pm$ 0.19 & \textbf{96.77 $\pm$ 0.13}\\
 & Gisette & 93.70 $\pm$ 0.00 & \textbf{97.34 $\pm$ 0.21} & 95.39 $\pm$ 0.60 & 97.10 $\pm$ 0.23\\
\midrule
 & Coil20 & 94.10 $\pm$ 0.00 & 95.03 $\pm$ 0.79 & 95.49 $\pm$ 1.10 & \textbf{98.02 $\pm$ 1.85}\\
 & ORL & 81.25 $\pm$ 0.00 & \textbf{92.00 $\pm$ 2.92} & 88.75 $\pm$ 3.06 & 89.50 $\pm$ 2.45\\
 & Yale & 72.73 $\pm$ 0.00 & \textbf{80.61 $\pm$ 2.78} & 67.88 $\pm$ 5.45 & 65.76 $\pm$ 7.05\\
\midrule
 & BASEHOCK & 87.72 $\pm$ 0.00 & 94.24 $\pm$ 0.59 & 82.66 $\pm$ 9.23 & \textbf{94.34 $\pm$ 0.66}\\
 & PCMAC & 66.07 $\pm$ 0.00 & \textbf{90.18 $\pm$ 0.99} & 87.12 $\pm$ 3.74 & 87.89 $\pm$ 1.26\\
 & RELATHE & 80.77 $\pm$ 0.00 & \textbf{83.64 $\pm$ 2.37} & 78.15 $\pm$ 5.32 & 82.10 $\pm$ 1.02\\
\midrule
 & Prostate\_GE & \textbf{90.48 $\pm$ 0.00} & 89.05 $\pm$ 2.18 & 89.52 $\pm$ 3.56 & 88.57 $\pm$ 2.33\\
 & SMK & 73.68 $\pm$ 0.00 & 69.74 $\pm$ 4.44 & 74.21 $\pm$ 8.39 & \textbf{80.53 $\pm$ 2.41}\\
 & CLL & 69.56 $\pm$ 0.00 & \textbf{90.87 $\pm$ 5.65} & 74.78 $\pm$ 8.65 & 78.26 $\pm$ 8.02\\
 & Carcinom & 85.71 $\pm$ 0.00 & \textbf{92.29 $\pm$ 1.83} & 88.29 $\pm$ 5.18 & 83.43 $\pm$ 5.83\\
 & Lymphoma & 65.00 $\pm$ 0.00 & \textbf{90.00 $\pm$ 2.67} & 81.00 $\pm$ 6.63 & 66.00 $\pm$ 7.35\\
\midrule
 & Arcene & 65.00 $\pm$ 0.00 & \textbf{71.25 $\pm$ 3.58} & 58.00 $\pm$ 7.14 & 67.75 $\pm$ 7.11\\
\midrule
 & HAR & 92.87 $\pm$ 0.00 & 94.68 $\pm$ 0.20 & 94.56 $\pm$ 0.39 & \textbf{94.90 $\pm$ 0.39}\\
\midrule
 & Isolet & 91.47 $\pm$ 0.00 & 92.30 $\pm$ 0.44 & 93.44 $\pm$ 0.26 & \textbf{95.53 $\pm$ 0.42}\\
\midrule
 & Madelon & 58.67 $\pm$ 0.00 & \textbf{77.12 $\pm$ 1.31} & 62.50 $\pm$ 6.34 & 76.63 $\pm$ 2.36\\
\midrule

            \end{tabular}
    }

 \end{table}

\looseness=-1
\textbf{Computational Costs Comparison.}
To compare the computational efficiency of these methods, we have computed the theoretical estimation of FLOPs count in Table \ref{tab:results_flops}. 

Examining the efficiency gains achieved by a sparse neural network in comparison to its dense counterpart commonly involves assessing the FLOPs (floating-point operations), as highlighted in prior works \cite{evci2020rigging, sokar2021dynamic}. The evaluation of \textit{FLOPs} serves to estimate the time complexity of an algorithm irrespective of its specific implementation. Given that current deep learning hardware is not tailored for sparse matrix computations, many methods for generating sparse neural networks resort to simulating sparsity through binary weight masks. Consequently, the execution time of such approaches may not accurately reflect their efficiency. Furthermore, the community is actively engaged in exploring dedicated sparse implementations for these networks \cite{hooker2021hardware}. In line with the theoretical nature of our paper, we defer consideration of these engineering aspects to future research. Therefore, our analysis of efficiency also incorporates FLOPs count.

We only show these numbers for neural network-based feature selection methods, as the computational efficiency of Lasso is very low, and it falls behind the performance of the competitors on most datasets. LassoNet sparsifies the input layer by dropping features during training; however, other layers of the network are dense. Moreover, due to the long training time and convergence, it needs much higher FLOPs than other methods. For example, it requires $88.2\%$ more FLOPs than SET-Attr. STG exploits a probabilistic gating mechanism in the input layer to select features; if a gate is converged to 0, it is removed from the network; if it is converged to 1, it is selected. when the gate probabilities are not converged to 0/1, a cutoff can be set on the gate/feature probabilities to select features. We observed in our experiments that while some gate probabilities might converge to small values, they do not converge to exact 0. As a result, all input neurons contribute to the network's output. Therefore, the model is dense during training and the FLOPs are much higher than SET-Attr in all cases considered.

\section{Analysis of Effects of Sparsity for Feature Selection}

\begin{table}[!b]
    \centering
    \caption{FLOPs comparison among neural network feature selection methods (All numbers are divided by $10^{12}$). \textbf{Bold} entries denote the best performers. }
    \vspace{2mm}
    \label{tab:results_flops}
    \centering
            \scalebox{1}{
             \begin{tabular}{r@{\hskip 0.07in}r@{\hskip 0.07in}r@{\hskip 0.07in}r@{\hskip 0.07in}r@{\hskip 0.07in}r@{\hskip 0.07in}r@{\hskip 0.07in}r@{\hskip 0.07in}r@{\hskip 0.07in}r@{\hskip 0.07in}r@{\hskip 0.07in}r@{\hskip 0.07in}r@{\hskip 0.07in}}
                \toprule

 \textbf{Dataset} & \textbf{LassoNet}  & \textbf{STG}  & \textbf{Dense-Attr} & \textbf{SET-Attr}\\\midrule

MNIST & 370.10 & 63.76 & 62.14 & \textbf{ 47.67 } & \\
USPS & 18.87 & 3.19 & 3.68 & \textbf{ 1.59 } & \\
Gisette & 72.29 & 36.73 & 18.75 & \textbf{ 8.85 } & \\\midrule
Coil20 & 4.04 & 1.56 & 1.07 & \textbf{ 0.55 } & \\
ORL & 0.95 & 0.43 & 0.40 & \textbf{ 0.36 } & \\
Yale & 0.55 & 0.18 & 0.13 & \textbf{ 0.10 } & \\\midrule
BASEHOCK & 30.76 & 9.49 & 4.25 & \textbf{ 0.24 } & \\
PCMAC & 28.09 & 6.32 & 3.05 & \textbf{ 0.14 } & \\
RELATHE & 33.53 & 6.06 & 3.02 & \textbf{ 0.16 } & \\\midrule
Prostate\_GE & 1.60 & 0.59 & 0.53 & \textbf{ 0.19 } & \\
SMK & 6.45 & 3.59 & 2.66 & \textbf{ 0.97 } & \\
CLL & 3.12 & 1.21 & 0.65 & \textbf{ 0.51 } & \\
Carcinom & 3.70 & 1.55 & 0.93 & \textbf{ 0.44 } & \\
Lymphoma & 1.26 & 0.38 & 0.22 & \textbf{ 0.12 } & \\\midrule
Arcene & 3.58 & 1.94 & 1.28 & \textbf{ 0.81 } & \\\midrule
HAR & 27.44 & 5.84 & 5.76 & \textbf{ 4.27 } & \\\midrule
Isolet & 20.72 & 5.39 & 4.15 & \textbf{ 0.32 } & \\\midrule
Madelon & 4.42 & 1.44 & 0.82 & \textbf{ 0.03 } & \\

\bottomrule
 
            \end{tabular}
    }

 \end{table}

\begin{figure*}[t!]
    \centering
    \vspace{-3mm}
    \begin{subfigure}[t]{0.48 \textwidth}
        \centering
        \includegraphics[width=\textwidth]{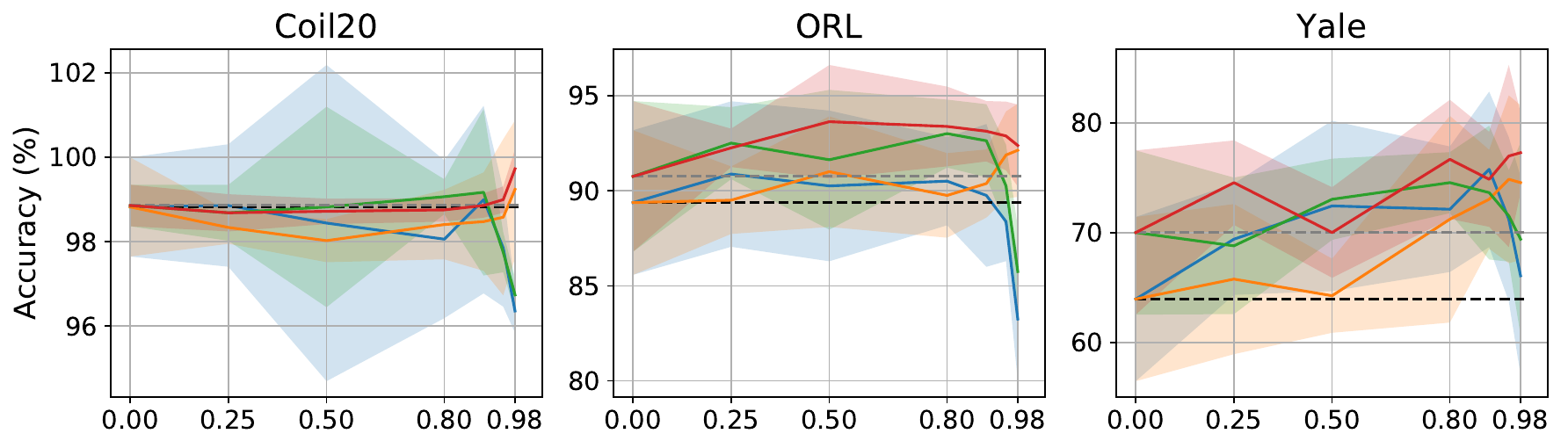}
        \caption{Face Image}\label{fig:categories_fs_image}
    \end{subfigure}
    \vspace{0mm}
    \begin{subfigure}[t]{0.48\textwidth}
        \centering
        \includegraphics[width=\textwidth]{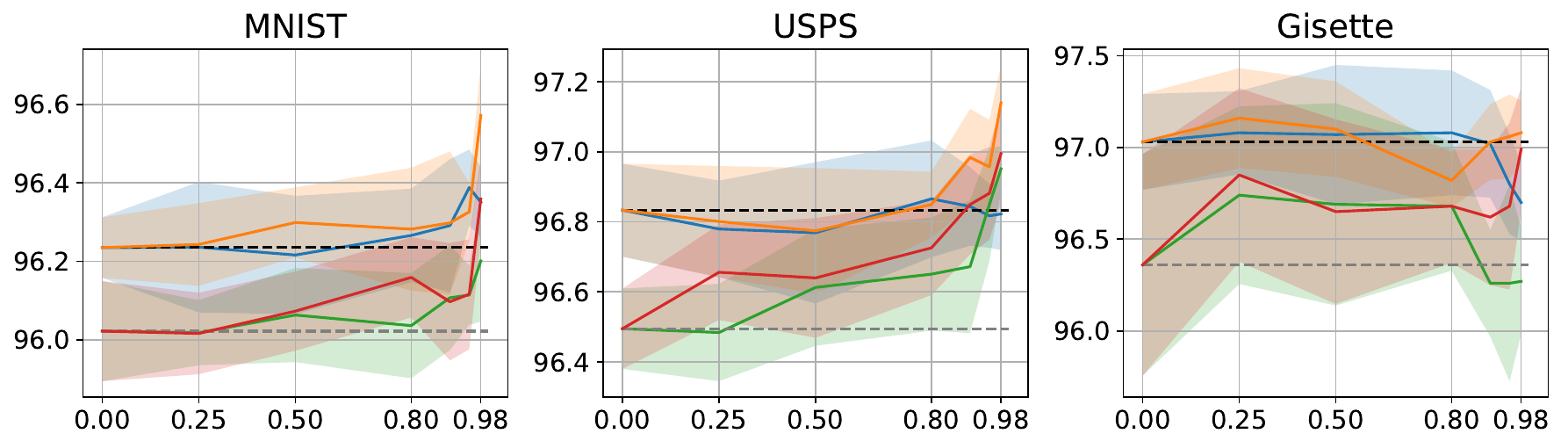}
        \caption{Digit Recognition}\label{fig:categories_fs_hand_written}
    \end{subfigure}%
    \vspace{0mm}
    \begin{subfigure}[t]{0.48\textwidth}
        \centering
        \includegraphics[width=\textwidth]{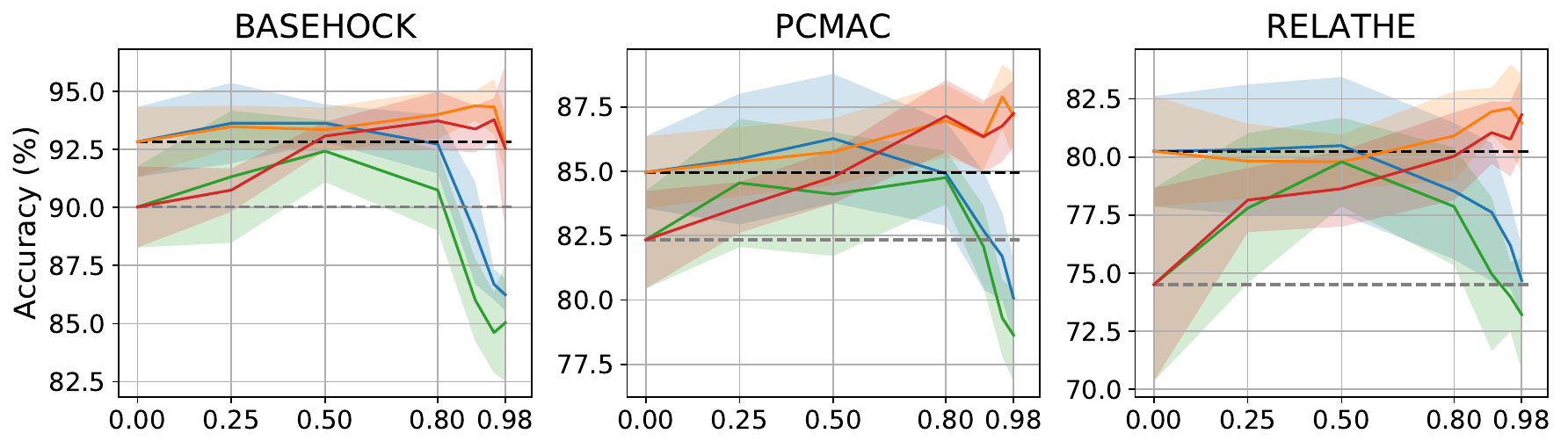}
        \caption{Text}\label{fig:categories_fs_text}
    \end{subfigure}%
     \vspace{0mm}
    \begin{subfigure}[t]{0.16\textwidth}
        \centering
        \includegraphics[width=\textwidth]{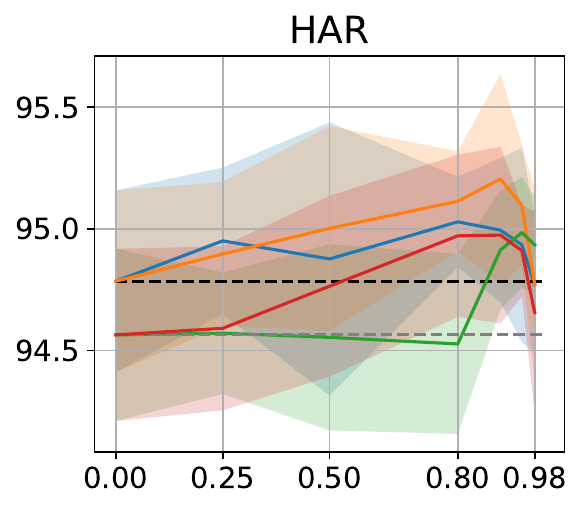}
        \caption{Time Series}\label{fig:categories_fs_time_series}
    \end{subfigure}%
    \vspace{0mm}
    \begin{subfigure}[t]{0.16\textwidth}
        \centering
        \includegraphics[width=\textwidth]{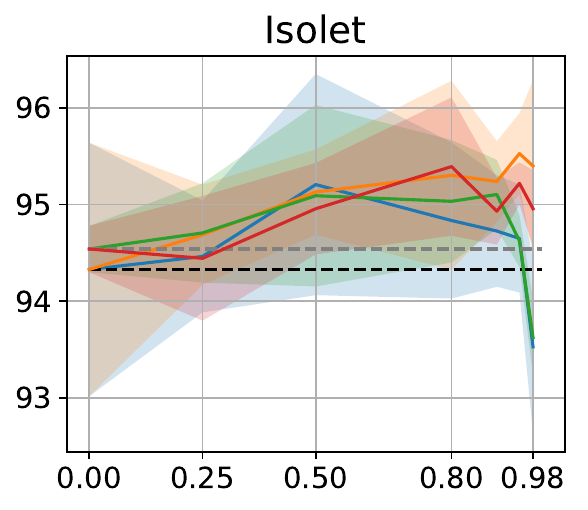}
        \caption{Speech}\label{fig:categories_fs_speech}
    \end{subfigure}%
    \vspace{0mm}
    \begin{subfigure}[t]{0.16\textwidth}
        \centering
        \includegraphics[width=\textwidth]{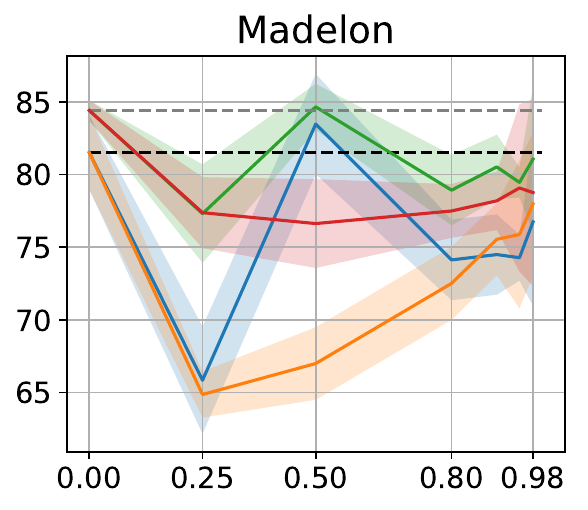}
        \caption{Noisy}\label{fig:categories_fs_noisy}
    \end{subfigure}%
    \vspace{0mm}
    \begin{subfigure}[t]{0.80\textwidth}
        \centering
        \includegraphics[width=\textwidth]{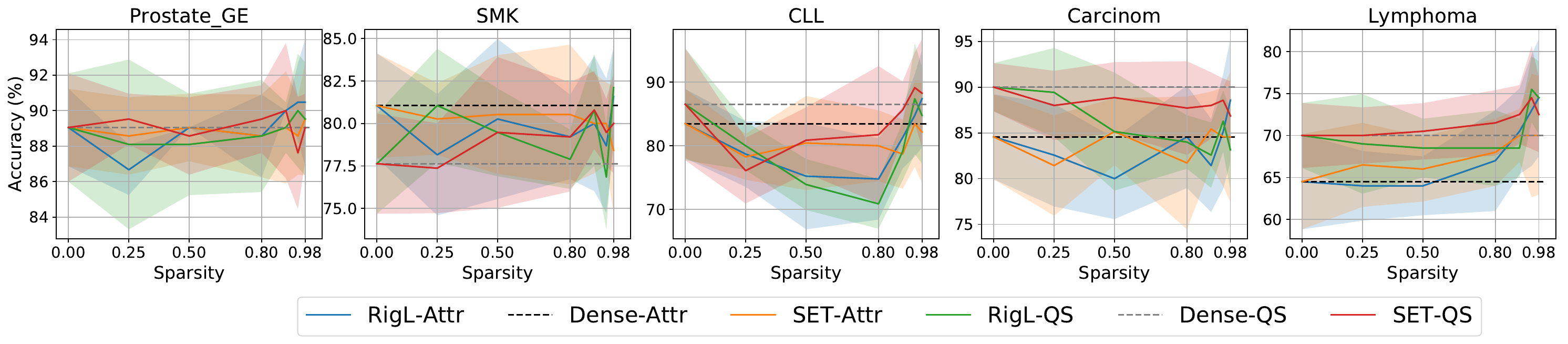}
        \caption{Biological}\label{fig:categories_fs_biological}
    \end{subfigure}%
    \vspace{0mm}
    \begin{subfigure}[t]{0.15\textwidth}
        \centering
        \includegraphics[width=\textwidth]{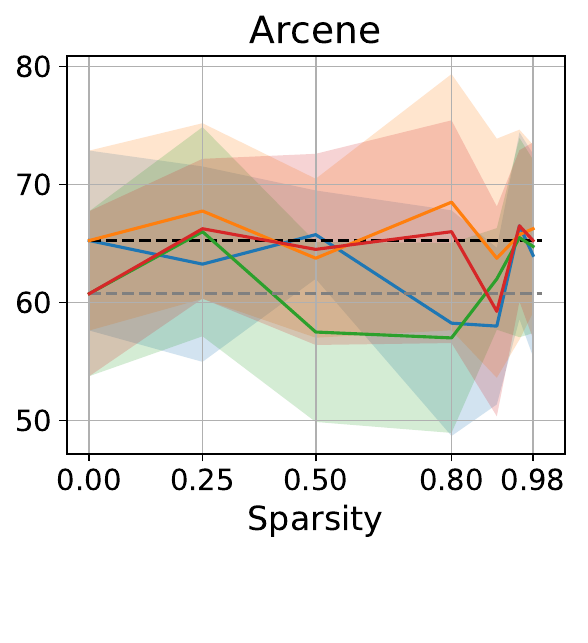}
        \caption{Arcene}\label{fig:categories_fs_arcene}
    \end{subfigure}%
    \vspace{2mm}
    \caption{Sparsity effect of feature selection performance.}\label{fig:categories_fs}
\end{figure*}

\textbf{Accuracy vs FLOPs Tradeoff.}
To summarize the results of Tables \ref{tab:results_fs_sparse} and \ref{tab:results_flops}, the tradeoff between accuracy and FLOPs is shown in Figure \ref{fig:acc_flops}. As we can observe, SET-Attr has in most cases in order of magnitude lower number of FLOPs than LassoNet while achieving competitive performance. STG and Dense-Attr fall behind in terms of both accuracy and FLOPs count in most cases considered.


\looseness=-1
In Section \ref{ssec:fs_100_comparison}, we optimize the sparsity level, selecting the value yielding minimal loss on the validation set. However, our focus now shifts to a comprehensive analysis of the impact of sparsity on feature selection performance, exploring a range of sparsity levels (i.e., ${0.25, 0.50, 0.75, 0.90, 0.95, 0.98}$) for each dataset. The results are visualized in Figure \ref{fig:categories_fs}.

\looseness=-1
Studying image datasets (Coil-20, ORL, Yale), we observe that sparsity levels up to $80\%$ generally enhance or maintain performance, with notable improvement on the Yale dataset. Beyond $80\%$, the feature selection method with SET (using any of the two importance metrics), remains stable or even exhibits improvement, whereas RigL experiences degradation in the high sparsity regime.

\looseness=-1
For hand-written digits datasets (MNIST, USPS, Gisette), sparsity consistently leads to performance improvement. Notably, the neuron attribution metric frequently outperforms neuron strength. At very high sparsity ($98\%$), feature selection with SET demonstrates significant performance gains, while RigL's performance either slightly improves or even declines.

\looseness=-1
On text datasets (BASEHOCK, PCMAC, REALTHE), performance improvement occurs until $50\%$ sparsity for all methods. However, RigL's feature selection performance, using both importance metrics, start to decline beyond $50\%$ sparsity, while SET continues to improve. Neuron attribution consistently outperforms neuron strength across most cases.

\looseness=-1
Across other high-dimensional datasets, the observed patterns vary. SET-QS often benefits from high sparsity ($>90\%$), outperforming dense networks in many cases. The Lymphoma dataset achieves maximum performance at $95\%$ sparsity for all methods. SMK, Prostate\_GE, Carcinom, and Arcene can be highly sparsified without significant performance drop, providing computational efficiency.

\looseness=-1
On HAR and Isolet datasets, sparsity enhances performance up to $90\%$ sparsity level. However, on the highly noisy Madelon dataset, sparsity generally deteriorates performance, with RigL-QS demonstrating superior performance among sparse models. Notably, RigL outperforms SET overall on the Madelon dataset.

This study emphasizes the importance of considering dataset characteristics when selecting an appropriate sparsity level for feature selection.



\section{Conclusions}
In conclusion, our work contributes significantly to the understanding of feature selection with sparse neural networks within the dynamic sparse training framework. Through systematic analysis, we demonstrate the efficacy of sparse neural networks in feature selection from diverse datasets, comparing them against dense neural networks and other sparsity-inducing methods. Our findings reveal that SNNs trained with DST algorithms achieve remarkable memory and computational savings, exceeding $50\%$ and $55\%$ respectively compared to dense networks, while maintaining superior feature selection quality in $13$ out of $18$ cases. One promising direction to continue this research is to consider neuron importance metrics to improve the training of sparse neural networks in the DST framework to guide the weight addition process.

\bibliography{resources}
\newpage
\appendix
\onecolumn
\section*{Appendices}

\section{Experimental Setup}\label{app:settings}
\subsection{Hyperparameters \& Model Architecture}
The architecture used in the experiments is an MLP with two hidden layers with 1000 and 100 hidden neurons in each layer, respectively. The hidden layer activation function is ReLU, and softmax is used for the output layer. We used a learning rate of $0.001$ for all datasets. The L2 regularization term and sparsity level are optimized based on the validation loss in $[5e^{-5},  0.0001, 0.001]$ and $[0.25, 0.5, 0.80, 0.9, 0.95, 0.98]$, respectively. $\zeta$ is set to $0.3$. All datasets have been scaled to have zero mean and unit variance. The batch size for the datasets with less than $200$ samples is set to $32$ and for the rest of the datasets is set to $100$. Adam optimizer is used for training the model. The maximum number of training epochs is $200$; if the validation loss does not improve within $50$ epochs, the training will end. The datasets are split into train, validation, and test sets with a split ratio $[65\%, 15\%, 20\%]$. The code is implemented in Python and using the PyTorch library \cite{Paszke_PyTorch_An_Imperative_2019}. The start of the code is \emph{GraNet}\footnote{\hyperlink{https://github.com/VITA-Group/GraNet}{https://github.com/VITA-Group/GraNet}} and \emph{TANGOS}\footnote{\hyperlink{https://github.com/alanjeffares/TANGOS}{https://github.com/alanjeffares/TANGOS}}.

\subsection{DST Algorithms}\label{s-app:settings:baselines}
In Section \ref{sssec:dst}, we explain the dynamic sparse training framework. In the experiments, we consider two dynamic sparse training algorithms: SET and RigL. These two algorithms are among the commonly used DST approaches in the literature to study the DST framework.  
\begin{itemize}
    \item \textbf{SET.} Sparse Evolutionary Training (SET) \cite{mocanu2018scalable} is a pioneering approach that introduced the Dynamic Sparse Training framework. This method starts with initializing a sparse network with the desired sparsity level. Then, at each epoch or every few iterations, it updates the connectivity of the sparse neural network by dropping a fraction $\zeta$ of weight with the smallest magnitude and adding the same number of random weights back to the network. This dynamic process serves as a means to continuously refine and update the network's topology, to evolve its structure over time.

    \item \textbf{RigL.} The Rigged Lottery (RigL) \cite{evci2020rigging} is another dynamic sparse training that evolves the topology of the sparse neural network by dynamically updating the connectivity. However, the difference compared to SET is that RigL adds the new weight based on the gradient information. It adds the non-existing weights having a large gradient magnitude to the network. 
\end{itemize}
\section{Performance for Various K Values}\label{app:K_values}
\begin{figure}[!h]
  \centering
  \includegraphics[width=0.35\textwidth]{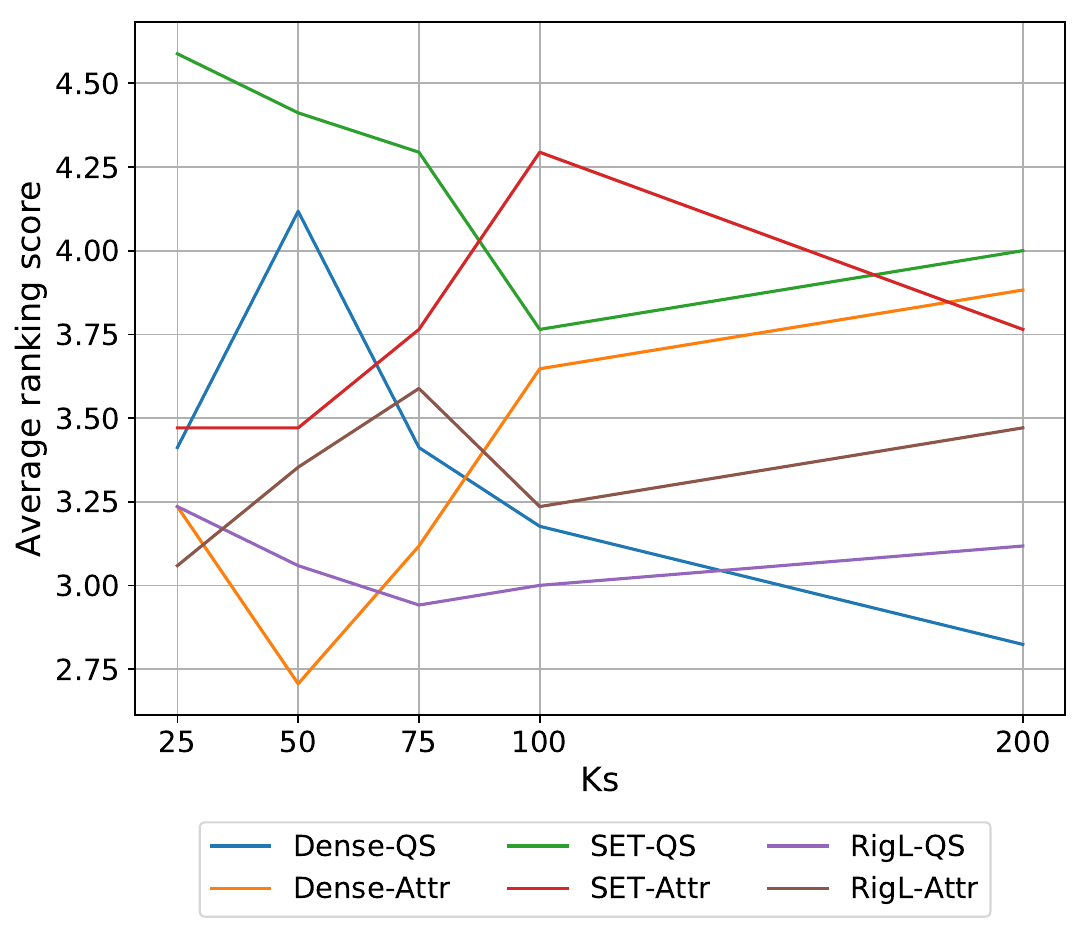} 
  \vspace{2mm}
  \caption{Average ranking score between the methods in Table \ref{tab:results_fs_all} for various K values.}
  \label{fig:avg_rank}
\end{figure}

While the results of the experiments in Section \ref{ssec:fs_100_comparison} are for when we select $K=100$ features, we measure the performance when selecting $K \in\{25, 50, 75, 100, 200\}$. To summarize the performance of each method when selecting different numbers of $K$, we compute the average ranking score. To compute this score, in each experiment (each dataset and value of $K$), we rank the methods in terms of accuracy and give a score depending on their rank, where the best-performing method receives a score of $6$ and the worst-performing method receives $1$. Then, we average these scores for all the experiments for each method. The average ranking score for each $K$ value is presented in Figure \ref{fig:avg_rank}.

\section{Experiment on Synthetic Dataset}\label{app:synthetic}
\begin{figure}[!h]
  \centering
  \includegraphics[width=0.6\textwidth]{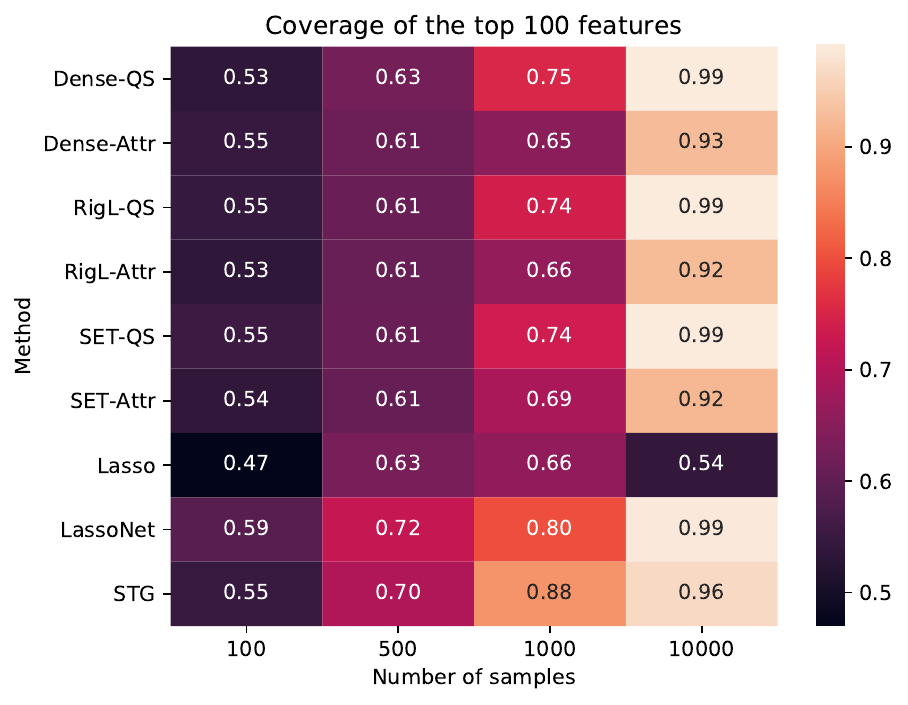} 
  \vspace{2mm}
  \caption{Coverage of the top 100 features for various methods considered in the experiments.}
  
  \label{fig:synthetic}
\end{figure}

To evaluate the performance of the methods in a controlled environment, we design an experiment on an artificial dataset. We generate an artificial dataset with 200 features where only 100 of these features are informative and the rest of the features are noise. The task is a binary classification. We vary the number of samples in $\{100, 500, 1000, 10000\}$. We plot the fraction of the relevant features that each method can find (coverage) in Figure \ref{fig:synthetic}.

As shown in Figure \ref{fig:synthetic}, as the number of samples increases the coverage also increases where the maximum value for most methods at 10000 samples. Neuron strength metric outperforms neuron attribution metric  for higher number of samples (1000 and 10000) in dense and sparse networks. We observed this also in some cases on the madelon dataset in Table \ref{tab:results_fs} in Section \ref{ssec:fs_100_comparison}. Lasso is the worst performers among all methods. LassoNet and STG outperform the other methods in medium size of samples (500 and 1000), while at the high number of samples LassoNet achieves on par performance with QuickSelection. We should also highlight again that the computational costs of sparse neural network-based methods are much fewer than the competitors as shown in Section \ref{ssec:sparsity_comparison}. However, the can achieve comparable performance to the competitors and find informative features.

\section{Feature Selection Comparison}\label{app:more_comparison}
The overall comparison among the methods in shown in Table \ref{tab:results_fs_all}. In addition, more pairwise comparison (as shown in Table \ref{tab:results_fs}), are brought in Table \ref{tab:results_fs_more}.
\begin{table*}[!h]
    \centering
    \caption{Feature selection results in terms of test classification accuracy [\%] of an SVM classifier on the selected subset of (K=100). The values in the parenthesis show the sparsity level.}
    \vspace{2mm}
    \label{tab:results_fs_all}
    \centering
            \scalebox{0.95}{
             \begin{tabular}{c@{\hskip 0.07in}c@{\hskip 0.07in}c@{\hskip 0.07in}c@{\hskip 0.07in}c@{\hskip 0.07in}c@{\hskip 0.07in}c@{\hskip 0.07in}c@{\hskip 0.07in}c@{\hskip 0.07in}c@{\hskip 0.07in}c@{\hskip 0.07in}c@{\hskip 0.07in}c@{\hskip 0.07in}}
                \toprule

& \textbf{Dataset}  & \textbf{Dense-QS} & \textbf{Dense-Attr} & \textbf{SET-QS} & \textbf{SET-Attr} & \textbf{RigL-QS} & \textbf{RigL-Attr}\\\midrule

Hand-written & MNIST & 96.02 $\pm$ 0.13 (0.00) & 96.23 $\pm$ 0.08 (0.00) & 96.02 $\pm$ 0.10 (0.25) & \textbf{96.24 $\pm$ 0.13} (0.25) & 96.06 $\pm$ 0.08 (0.50) & 96.22 $\pm$ 0.10 (0.50)\\   
 & USPS & 96.49 $\pm$ 0.11 (0.00) & 96.83 $\pm$ 0.13 (0.00) & 96.64 $\pm$ 0.13 (0.50) & 96.77 $\pm$ 0.13 (0.50) & 96.65 $\pm$ 0.16 (0.80) & \textbf{96.87 $\pm$ 0.17} (0.80)\\
 & Gisette & 96.36 $\pm$ 0.60 (0.00) & 97.03 $\pm$ 0.26 (0.00) & 96.65 $\pm$ 0.45 (0.50) & \textbf{97.10 $\pm$ 0.23} (0.50) & 96.69 $\pm$ 0.53 (0.50) & 97.07 $\pm$ 0.27 (0.50)\\
\midrule
Image & Coil20 & 98.85 $\pm$ 0.49 (0.00) & 98.82 $\pm$ 1.17 (0.00) & 98.72 $\pm$ 0.31 (0.50) & 98.02 $\pm$ 1.85 (0.50) & \textbf{99.06 $\pm$ 0.41} (0.80) & 98.06 $\pm$ 1.87 (0.80)\\
 & ORL & 90.75 $\pm$ 3.96 (0.00) & 89.38 $\pm$ 3.80 (0.00) & 92.25 $\pm$ 2.15 (0.25) & 89.50 $\pm$ 2.45 (0.25) & \textbf{92.50 $\pm$ 2.30} (0.25) & 90.88 $\pm$ 3.26 (0.25)\\
 & Yale & 70.00 $\pm$ 7.48 (0.00) & 63.94 $\pm$ 7.48 (0.00) & \textbf{74.55 $\pm$ 3.64} (0.25) & 65.76 $\pm$ 7.05 (0.25) & 71.52 $\pm$ 5.28 (0.50) & 73.94 $\pm$ 5.94 (0.50)\\
\midrule
Text & BASEHOCK & 90.03 $\pm$ 1.75 (0.00) & 92.83 $\pm$ 1.51 (0.00) & 93.78 $\pm$ 1.01 (0.95) & \textbf{94.34 $\pm$ 0.66} (0.95) & 84.61 $\pm$ 1.77 (0.95) & 86.69 $\pm$ 2.20 (0.95)\\        
 & PCMAC & 82.34 $\pm$ 1.90 (0.00) & 84.96 $\pm$ 1.40 (0.00) & 86.76 $\pm$ 1.41 (0.95) & \textbf{87.89 $\pm$ 1.26} (0.95) & 79.31 $\pm$ 1.59 (0.95) & 81.70 $\pm$ 2.29 (0.95)\\
 & RELATHE & 74.51 $\pm$ 4.16 (0.00) & 80.24 $\pm$ 2.38 (0.00) & 80.77 $\pm$ 1.35 (0.95) & \textbf{82.10 $\pm$ 1.02} (0.95) & 73.22 $\pm$ 1.90 (0.98) & 74.69 $\pm$ 2.96 (0.98)\\
\midrule
Biological & Prostate\_GE & \textbf{89.05 $\pm$ 3.05} (0.00) & \textbf{89.05 $\pm$ 2.18} (0.00) & 87.62 $\pm$ 4.86 (0.50) & 88.57 $\pm$ 2.33 (0.50) & 85.71 $\pm$ 5.63 (0.80) & 88.09 $\pm$ 3.19 (0.80)\\
 & SMK & 77.63 $\pm$ 2.94 (0.00) & \textbf{81.05 $\pm$ 3.07} (0.00) & 79.47 $\pm$ 1.97 (0.50) & 80.53 $\pm$ 2.41 (0.50) & 79.47 $\pm$ 3.07 (0.50) & 80.26 $\pm$ 3.95 (0.50)\\
 & CLL & \textbf{86.52 $\pm$ 8.79} (0.00) & 83.48 $\pm$ 5.43 (0.00) & 76.09 $\pm$ 8.53 (0.25) & 78.26 $\pm$ 8.02 (0.25) & 80.00 $\pm$ 7.58 (0.25) & 78.70 $\pm$ 7.64 (0.25)\\
 & Carcinom & \textbf{90.00 $\pm$ 2.63} (0.00) & 84.57 $\pm$ 4.64 (0.00) & 89.14 $\pm$ 2.49 (0.50) & 83.43 $\pm$ 5.83 (0.50) & 85.14 $\pm$ 3.33 (0.50) & 80.00 $\pm$ 6.26 (0.50)\\
 & Lymphoma & 70.00 $\pm$ 3.87 (0.00) & 64.50 $\pm$ 5.68 (0.00) & \textbf{70.50 $\pm$ 6.10} (0.50) & 66.00 $\pm$ 7.35 (0.50) & 67.00 $\pm$ 4.58 (0.50) & 65.50 $\pm$ 7.57 (0.50)\\
\midrule
Mass Spectrometry & Arcene & 60.75 $\pm$ 6.99 (0.00) & 65.25 $\pm$ 7.62 (0.00) & 66.25 $\pm$ 8.31 (0.25) & \textbf{67.75 $\pm$ 7.11} (0.25) & 58.00 $\pm$ 6.00 (0.80) & 57.00 $\pm$ 6.40 (0.80)\\
\midrule
Time-series & HAR & 94.56 $\pm$ 0.35 (0.00) & 94.78 $\pm$ 0.37 (0.00) & 94.59 $\pm$ 0.41 (0.25) & \textbf{94.90 $\pm$ 0.39} (0.25) & 94.55 $\pm$ 0.23 (0.50) & 94.88 $\pm$ 0.40 (0.50)\\      
\midrule
Speech & Isolet & 94.54 $\pm$ 0.24 (0.00) & 94.33 $\pm$ 1.31 (0.00) & 95.22 $\pm$ 0.35 (0.95) & \textbf{95.53 $\pm$ 0.42} (0.95) & 95.10 $\pm$ 0.51 (0.90) & 94.72 $\pm$ 0.58 (0.90)\\        
\midrule
Noisy & Madelon & \textbf{84.42 $\pm$ 0.69} (0.00) & 81.52 $\pm$ 2.55 (0.00) & 78.40 $\pm$ 1.71 (0.95) & 76.63 $\pm$ 2.36 (0.95) & 79.80 $\pm$ 1.99 (0.98) & 75.75 $\pm$ 3.50 (0.98)\\        
\midrule

            \end{tabular}
    }
 \end{table*}

\begin{table*}[!h]
\caption{Feature selection results in terms of test classification accuracy [\%] of an SVM classifier on the selected subset of (K=100). The values in the parenthesis show the sparsity level.}
\vspace{2mm}
\begin{subtable}[h]{0.3\textwidth}
        \centering
        \scalebox{1}{
        \begin{tabular}{c@{\hskip 0.07in}c@{\hskip 0.07in}c@{\hskip 0.07in}c}
        \toprule

            & \textbf{Dense-QS} & \textbf{Dense-Attr}\\\midrule

MNIST & 96.02 $\pm$ 0.13 (0.00) & \textbf{96.23 $\pm$ 0.08} (0.00)\\
USPS & 96.49 $\pm$ 0.11 (0.00) & \textbf{96.83 $\pm$ 0.13} (0.00)\\
Gisette & 96.36 $\pm$ 0.60 (0.00) & \textbf{97.03 $\pm$ 0.26} (0.00)\\
\midrule
Coil20 & \textbf{98.85 $\pm$ 0.49} (0.00) & 98.82 $\pm$ 1.17 (0.00)\\
ORL & \textbf{90.75 $\pm$ 3.96} (0.00) & 89.38 $\pm$ 3.80 (0.00)\\
Yale & \textbf{70.00 $\pm$ 7.48} (0.00) & 63.94 $\pm$ 7.48 (0.00)\\
\midrule
BASEHOCK & 90.03 $\pm$ 1.75 (0.00) & \textbf{92.83 $\pm$ 1.51} (0.00)\\
PCMAC & 82.34 $\pm$ 1.90 (0.00) & \textbf{84.96 $\pm$ 1.40} (0.00)\\
RELATHE & 74.51 $\pm$ 4.16 (0.00) & \textbf{80.24 $\pm$ 2.38} (0.00)\\
\midrule
Prostate\_GE & \textbf{89.05 $\pm$ 3.05} (0.00) & \textbf{89.05 $\pm$ 2.18} (0.00)\\
SMK & 77.63 $\pm$ 2.94 (0.00) & \textbf{81.05 $\pm$ 3.07} (0.00)\\
CLL & \textbf{86.52 $\pm$ 8.79} (0.00) & 83.48 $\pm$ 5.43 (0.00)\\
Carcinom & \textbf{90.00 $\pm$ 2.63} (0.00) & 84.57 $\pm$ 4.64 (0.00)\\
Lymphoma & \textbf{70.00 $\pm$ 3.87} (0.00) & 64.50 $\pm$ 5.68 (0.00)\\
\midrule
Arcene & 60.75 $\pm$ 6.99 (0.00) & \textbf{65.25 $\pm$ 7.62} (0.00)\\
\midrule
HAR & 94.56 $\pm$ 0.35 (0.00) & \textbf{94.78 $\pm$ 0.37} (0.00)\\
\midrule
Isolet & \textbf{94.54 $\pm$ 0.24} (0.00) & 94.33 $\pm$ 1.31 (0.00)\\
\midrule
Madelon & \textbf{84.42 $\pm$ 0.69} (0.00) & 81.52 $\pm$ 2.55 (0.00)\\
\midrule

       \end{tabular}}
       \caption{Dense-QS vs. Dense-Attr.}
       \label{tab:app:dense-qsVSattr}
    \end{subtable}
    \hfill\hfill\hfill\hfill\hfill\hfill\hfill\hfill\hfill\hfill\hfill\hfill\hfill
    \begin{subtable}[h]{0.3\textwidth}
        \centering
        \scalebox{1}{
        \begin{tabular}{c@{\hskip 0.0in}c@{\hskip 0.07in}c@{\hskip 0.07in}c}
        \toprule
            & \textbf{SET-QS} & \textbf{RigL-QS} \\\midrule

 & 96.02 $\pm$ 0.10 (0.25) & \textbf{96.06 $\pm$ 0.08} (0.50)\\
 & 96.64 $\pm$ 0.13 (0.50) & \textbf{96.65 $\pm$ 0.16} (0.80)\\
 & 96.65 $\pm$ 0.45 (0.50) & \textbf{96.69 $\pm$ 0.53} (0.50)\\
\midrule
 & 98.72 $\pm$ 0.31 (0.50) & \textbf{99.06 $\pm$ 0.41} (0.80)\\
 & 92.25 $\pm$ 2.15 (0.25) & \textbf{92.50 $\pm$ 2.30} (0.25)\\
 & \textbf{74.55 $\pm$ 3.64} (0.25) & 71.52 $\pm$ 5.28 (0.50)\\
\midrule
 & \textbf{93.78 $\pm$ 1.01} (0.95) & 84.61 $\pm$ 1.77 (0.95)\\
 & \textbf{86.76 $\pm$ 1.41} (0.95) & 79.31 $\pm$ 1.59 (0.95)\\
 & \textbf{80.77 $\pm$ 1.35} (0.95) & 73.22 $\pm$ 1.90 (0.98)\\
\midrule
 & \textbf{87.62 $\pm$ 4.86} (0.50) & 85.71 $\pm$ 5.63 (0.80)\\
 & \textbf{79.47 $\pm$ 1.97} (0.50) & \textbf{79.47 $\pm$ 3.07} (0.50)\\
 & 76.09 $\pm$ 8.53 (0.25) & \textbf{80.00 $\pm$ 7.58} (0.25)\\
 & \textbf{89.14 $\pm$ 2.49} (0.50) & 85.14 $\pm$ 3.33 (0.50)\\
 & \textbf{70.50 $\pm$ 6.10} (0.50) & 67.00 $\pm$ 4.58 (0.50)\\
\midrule
 & \textbf{66.25 $\pm$ 8.31} (0.25) & 58.00 $\pm$ 6.00 (0.80)\\
\midrule
 & \textbf{94.59 $\pm$ 0.41} (0.25) & 94.55 $\pm$ 0.23 (0.50)\\
\midrule
 & \textbf{95.22 $\pm$ 0.35} (0.95) & 95.10 $\pm$ 0.51 (0.90)\\
\midrule
 & 78.40 $\pm$ 1.71 (0.95) & \textbf{79.80 $\pm$ 1.99} (0.98)\\
\midrule

       \end{tabular}}
       \caption{RigL-QS vs. SET-QS.}
       \label{tab:app:QS-SETvsRigL}
    \end{subtable}
    \hfill
    \begin{subtable}[h]{0.3\textwidth}
        \centering
        \scalebox{1}{
        \begin{tabular}{c@{\hskip 0.0in}c@{\hskip 0.07in}c@{\hskip 0.07in}c}
        \toprule

            & \textbf{RigL-Attr} & \textbf{RigL-QS}\\\midrule

 & \textbf{96.22 $\pm$ 0.10} (0.50) & 96.06 $\pm$ 0.08 (0.50)\\
 & \textbf{96.87 $\pm$ 0.17} (0.80) & 96.65 $\pm$ 0.16 (0.80)\\
 & \textbf{97.07 $\pm$ 0.27} (0.50) & 96.69 $\pm$ 0.53 (0.50)\\
\midrule
 & 98.06 $\pm$ 1.87 (0.80) & \textbf{99.06 $\pm$ 0.41} (0.80)\\
 & 90.88 $\pm$ 3.26 (0.25) & \textbf{92.50 $\pm$ 2.30} (0.25)\\
 & \textbf{73.94 $\pm$ 5.94} (0.50) & 71.52 $\pm$ 5.28 (0.50)\\
\midrule
 & \textbf{86.69 $\pm$ 2.20} (0.95) & 84.61 $\pm$ 1.77 (0.95)\\
 & \textbf{81.70 $\pm$ 2.29} (0.95) & 79.31 $\pm$ 1.59 (0.95)\\
 & \textbf{74.69 $\pm$ 2.96} (0.98) & 73.22 $\pm$ 1.90 (0.98)\\
\midrule
 & \textbf{88.09 $\pm$ 3.19} (0.80) & 85.71 $\pm$ 5.63 (0.80)\\
 & \textbf{80.26 $\pm$ 3.95} (0.50) & 79.47 $\pm$ 3.07 (0.50)\\
 & 78.70 $\pm$ 7.64 (0.25) & \textbf{80.00 $\pm$ 7.58} (0.25)\\
 & 80.00 $\pm$ 6.26 (0.50) & \textbf{85.14 $\pm$ 3.33} (0.50)\\
 & 65.50 $\pm$ 7.57 (0.50) & \textbf{67.00 $\pm$ 4.58} (0.50)\\
\midrule
 & 57.00 $\pm$ 6.40 (0.80) & \textbf{58.00 $\pm$ 6.00} (0.80)\\
\midrule
 & \textbf{94.88 $\pm$ 0.40} (0.50) & 94.55 $\pm$ 0.23 (0.50)\\
\midrule
 & 94.72 $\pm$ 0.58 (0.90) & \textbf{95.10 $\pm$ 0.51} (0.90)\\
\midrule
 & 75.75 $\pm$ 3.50 (0.98) & \textbf{79.80 $\pm$ 1.99} (0.98)\\
\midrule
       \end{tabular}}
       \caption{SET-QS vs. SET-Attr.}
       \label{tab:app:RigL-QSvsAttr}
    \end{subtable}
    \hfill

     \label{tab:results_fs_more}
\end{table*}

\section{Feature Importance Metric Analysis}\label{app:importance_metric}
In the experiments in the paper, we calculate the neuron importance based on the summation of the importance within all training epochs. However, we considered two other ways to compute the importance: the summation of importance within last training epoch, and the importance in the last iteration.  The results of these two approaches are summarized in Tables \ref{tab:results_fs_epoch} and \ref{tab:results_fs_itr}. As can be seen from this Table, considering the importance within all training epochs, as done in Section \ref{ssec:fs_100_comparison} leads to better results. This shows that the history of the importance, even when measured at the first epochs is also beneficial to the performance.
\begin{table*}[!h]
    \centering
    \caption{Feature selection results in terms of test classification accuracy [\%] of an SVM classifier on the selected subset of (K=100). The importance metric is computed based on the summation of importance values (for all iterations) within the last training epoch for all methods. The values in the parenthesis show the sparsity level.}
    \vspace{2mm}
    \label{tab:results_fs_epoch}
    \centering
            \scalebox{0.95}{
             \begin{tabular}{c@{\hskip 0.07in}c@{\hskip 0.07in}c@{\hskip 0.07in}c@{\hskip 0.07in}c@{\hskip 0.07in}c@{\hskip 0.07in}c@{\hskip 0.07in}c@{\hskip 0.07in}c@{\hskip 0.07in}c@{\hskip 0.07in}c@{\hskip 0.07in}c@{\hskip 0.07in}c@{\hskip 0.07in}}
                \toprule

& \textbf{Dataset}  & \textbf{Dense-QS} & \textbf{Dense-Attr} & \textbf{SET-QS} & \textbf{SET-Attr} & \textbf{RigL-QS} & \textbf{RigL-Attr}\\\midrule

Hand-written & MNIST & 95.97 $\pm$ 0.24 (0.00) & 96.21 $\pm$ 0.17 (0.00) & 95.89 $\pm$ 0.22 (0.25) & \textbf{96.23 $\pm$ 0.17} (0.25) & 95.90 $\pm$ 0.21 (0.50) & 96.19 $\pm$ 0.20 (0.50)\\   
 & USPS & 96.60 $\pm$ 0.20 (0.00) & 96.81 $\pm$ 0.19 (0.00) & 96.67 $\pm$ 0.26 (0.50) & \textbf{96.91 $\pm$ 0.12} (0.50) & 96.58 $\pm$ 0.14 (0.80) & 96.86 $\pm$ 0.14 (0.80)\\
 & Gisette & 94.52 $\pm$ 1.26 (0.00) & 96.66 $\pm$ 0.40 (0.00) & 93.91 $\pm$ 1.79 (0.50) & \textbf{96.99 $\pm$ 0.40} (0.50) & 95.40 $\pm$ 0.60 (0.50) & 96.80 $\pm$ 0.36 (0.50)\\
\midrule
Image & Coil20 & 98.82 $\pm$ 0.59 (0.00) & 98.82 $\pm$ 1.20 (0.00) & 98.75 $\pm$ 0.32 (0.50) & 98.75 $\pm$ 0.35 (0.50) & \textbf{99.03 $\pm$ 0.43} (0.80) & 98.78 $\pm$ 0.61 (0.80)\\
 & ORL & \textbf{90.88 $\pm$ 2.56} (0.00) & 89.25 $\pm$ 3.92 (0.00) & 90.62 $\pm$ 2.32 (0.25) & 88.62 $\pm$ 2.12 (0.25) & 90.75 $\pm$ 1.79 (0.25) & 90.38 $\pm$ 3.40 (0.25)\\
 & Yale & 68.48 $\pm$ 5.94 (0.00) & 59.39 $\pm$ 8.48 (0.00) & \textbf{72.73 $\pm$ 7.17} (0.25) & 63.33 $\pm$ 7.95 (0.25) & 68.49 $\pm$ 5.94 (0.50) & 71.82 $\pm$ 6.78 (0.50)\\
\midrule
Text & BASEHOCK & 75.59 $\pm$ 6.68 (0.00) & 86.69 $\pm$ 2.44 (0.00) & 92.00 $\pm$ 1.60 (0.95) & 90.85 $\pm$ 1.84 (0.95) & 87.84 $\pm$ 1.95 (0.95) & \textbf{92.06 $\pm$ 1.81} (0.95)\\        
 & PCMAC & 78.20 $\pm$ 2.98 (0.00) & 80.69 $\pm$ 1.28 (0.00) & \textbf{85.50 $\pm$ 1.21} (0.95) & 84.73 $\pm$ 1.47 (0.95) & 81.65 $\pm$ 2.24 (0.95) & 84.04 $\pm$ 2.05 (0.95)\\
 & RELATHE & 64.97 $\pm$ 6.84 (0.00) & 73.53 $\pm$ 3.14 (0.00) & \textbf{80.42 $\pm$ 1.71} (0.95) & 78.57 $\pm$ 2.97 (0.95) & 73.36 $\pm$ 1.83 (0.98) & 78.08 $\pm$ 2.62 (0.98)\\
\midrule
Biological & Prostate\_GE & 89.05 $\pm$ 4.29 (0.00) & \textbf{90.00 $\pm$ 1.43} (0.00) & 89.52 $\pm$ 1.90 (0.50) & 88.09 $\pm$ 3.19 (0.50) & 87.62 $\pm$ 3.16 (0.80) & 87.62 $\pm$ 2.33 (0.80)\\
 & SMK & 75.53 $\pm$ 4.25 (0.00) & 79.74 $\pm$ 3.13 (0.00) & 80.53 $\pm$ 4.11 (0.50) & \textbf{81.05 $\pm$ 3.29} (0.50) & 78.95 $\pm$ 4.24 (0.50) & 80.26 $\pm$ 5.16 (0.50)\\
 & CLL & \textbf{79.57 $\pm$ 14.43} (0.00) & 77.39 $\pm$ 7.97 (0.00) & 73.91 $\pm$ 16.38 (0.25) & 74.78 $\pm$ 13.72 (0.25) & 71.74 $\pm$ 7.84 (0.25) & 72.61 $\pm$ 7.79 (0.25)\\
 & Carcinom & \textbf{84.57 $\pm$ 5.74} (0.00) & 77.14 $\pm$ 7.23 (0.00) & \textbf{84.57 $\pm$ 6.29} (0.50) & 75.71 $\pm$ 4.09 (0.50) & 80.57 $\pm$ 6.49 (0.50) & 78.29 $\pm$ 5.14 (0.50)\\   
 & Lymphoma & 70.00 $\pm$ 5.00 (0.00) & 64.00 $\pm$ 4.36 (0.00) & \textbf{71.50 $\pm$ 7.43} (0.50) & 63.00 $\pm$ 5.57 (0.50) & 65.50 $\pm$ 4.15 (0.50) & 63.00 $\pm$ 6.00 (0.50)\\
\midrule
Mass Spectrometry & Arcene & 65.50 $\pm$ 9.07 (0.00) & \textbf{65.75 $\pm$ 7.59} (0.00) & 63.50 $\pm$ 8.08 (0.25) & 63.75 $\pm$ 6.64 (0.25) & 56.75 $\pm$ 5.60 (0.80) & 63.25 $\pm$ 5.60 (0.80)\\
\midrule
Time-series & HAR & 94.72 $\pm$ 0.30 (0.00) & 94.85 $\pm$ 0.25 (0.00) & 94.30 $\pm$ 0.33 (0.25) & \textbf{94.88 $\pm$ 0.30} (0.25) & 94.64 $\pm$ 0.33 (0.50) & 94.63 $\pm$ 0.34 (0.50)\\      
\midrule
Speech & Isolet & 94.11 $\pm$ 0.46 (0.00) & 94.23 $\pm$ 0.69 (0.00) & 95.57 $\pm$ 0.21 (0.95) & \textbf{95.64 $\pm$ 0.25} (0.95) & 95.04 $\pm$ 0.42 (0.90) & 95.01 $\pm$ 0.42 (0.90)\\        
\midrule
Noisy & Madelon & \textbf{84.30 $\pm$ 0.96} (0.00) & 82.25 $\pm$ 2.20 (0.00) & 78.45 $\pm$ 2.44 (0.95) & 75.87 $\pm$ 2.94 (0.95) & 79.13 $\pm$ 2.20 (0.98) & 72.30 $\pm$ 3.86 (0.98)\\        
\midrule

            \end{tabular}
    }

 \end{table*} 
\begin{table*}[!h]
    \centering
    \caption{Feature selection results in terms of test classification accuracy [\%] of an SVM classifier on the selected subset of (K=100). The importance metric is computed at the last training iteration for all methods. The values in the parenthesis show the sparsity level.}
    \vspace{2mm}
    \label{tab:results_fs_itr}
    \centering
            \scalebox{0.95}{
             \begin{tabular}{c@{\hskip 0.07in}c@{\hskip 0.07in}c@{\hskip 0.07in}c@{\hskip 0.07in}c@{\hskip 0.07in}c@{\hskip 0.07in}c@{\hskip 0.07in}c@{\hskip 0.07in}c@{\hskip 0.07in}c@{\hskip 0.07in}c@{\hskip 0.07in}c@{\hskip 0.07in}c@{\hskip 0.07in}}
                \toprule

& \textbf{Dataset}  & \textbf{Dense-QS} & \textbf{Dense-Attr} & \textbf{SET-QS} & \textbf{SET-Attr} & \textbf{RigL-QS} & \textbf{RigL-Attr}\\\midrule

Hand-written & MNIST & 96.10 $\pm$ 0.09 (0.00) & 96.13 $\pm$ 0.20 (0.00) & 95.97 $\pm$ 0.22 (0.25) & 96.06 $\pm$ 0.20 (0.25) & 95.99 $\pm$ 0.20 (0.50) & \textbf{96.16 $\pm$ 0.34} (0.50)\\   
 & USPS & 96.55 $\pm$ 0.24 (0.00) & \textbf{96.81 $\pm$ 0.24} (0.00) & 96.60 $\pm$ 0.17 (0.50) & 96.76 $\pm$ 0.19 (0.50) & 96.58 $\pm$ 0.19 (0.80) & 96.77 $\pm$ 0.18 (0.80)\\
 & Gisette & 93.36 $\pm$ 2.76 (0.00) & 96.70 $\pm$ 0.71 (0.00) & 94.37 $\pm$ 1.54 (0.50) & 96.82 $\pm$ 0.37 (0.50) & 95.17 $\pm$ 0.76 (0.50) & \textbf{96.86 $\pm$ 0.52} (0.50)\\
\midrule
Image & Coil20 & 98.82 $\pm$ 0.59 (0.00) & 98.61 $\pm$ 1.51 (0.00) & 98.78 $\pm$ 0.28 (0.50) & 98.68 $\pm$ 0.43 (0.50) & \textbf{99.03 $\pm$ 0.43} (0.80) & 98.78 $\pm$ 0.68 (0.80)\\
 & ORL & \textbf{91.00 $\pm$ 2.61} (0.00) & 87.75 $\pm$ 4.36 (0.00) & 90.75 $\pm$ 2.18 (0.25) & 90.00 $\pm$ 2.17 (0.25) & 90.75 $\pm$ 1.79 (0.25) & 89.00 $\pm$ 3.00 (0.25)\\
 & Yale & 68.48 $\pm$ 5.62 (0.00) & 63.64 $\pm$ 7.30 (0.00) & \textbf{71.82 $\pm$ 7.55} (0.25) & 61.82 $\pm$ 9.79 (0.25) & 68.18 $\pm$ 5.63 (0.50) & 65.15 $\pm$ 7.81 (0.50)\\
\midrule
Text & BASEHOCK & 74.26 $\pm$ 7.06 (0.00) & 85.31 $\pm$ 2.63 (0.00) & \textbf{91.80 $\pm$ 1.53} (0.95) & 90.20 $\pm$ 1.78 (0.95) & 88.22 $\pm$ 2.34 (0.95) & 91.13 $\pm$ 2.32 (0.95)\\        
 & PCMAC & 77.22 $\pm$ 4.17 (0.00) & 80.57 $\pm$ 1.17 (0.00) & \textbf{85.06 $\pm$ 1.65} (0.95) & 84.55 $\pm$ 1.38 (0.95) & 82.29 $\pm$ 1.61 (0.95) & 83.73 $\pm$ 2.08 (0.95)\\
 & RELATHE & 66.64 $\pm$ 4.95 (0.00) & 72.87 $\pm$ 2.46 (0.00) & \textbf{80.10 $\pm$ 1.97} (0.95) & 78.74 $\pm$ 3.62 (0.95) & 72.69 $\pm$ 2.39 (0.98) & 78.15 $\pm$ 2.50 (0.98)\\
\midrule
Biological & Prostate\_GE & 89.05 $\pm$ 4.29 (0.00) & \textbf{90.00 $\pm$ 1.43} (0.00) & 89.52 $\pm$ 1.90 (0.50) & 89.05 $\pm$ 3.05 (0.50) & 87.14 $\pm$ 3.05 (0.80) & 86.67 $\pm$ 3.56 (0.80)\\
 & SMK & 76.32 $\pm$ 2.88 (0.00) & 78.68 $\pm$ 3.62 (0.00) & \textbf{81.32 $\pm$ 4.47} (0.50) & 80.26 $\pm$ 3.95 (0.50) & 79.21 $\pm$ 3.62 (0.50) & 80.26 $\pm$ 4.74 (0.50)\\
 & CLL & \textbf{79.57 $\pm$ 11.35} (0.00) & 70.43 $\pm$ 8.65 (0.00) & 77.39 $\pm$ 14.39 (0.25) & 71.30 $\pm$ 13.07 (0.25) & 72.61 $\pm$ 7.54 (0.25) & 71.74 $\pm$ 12.34 (0.25)\\
 & Carcinom & 84.00 $\pm$ 6.41 (0.00) & 76.00 $\pm$ 6.54 (0.00) & \textbf{84.86 $\pm$ 6.13} (0.50) & 73.14 $\pm$ 3.88 (0.50) & 81.43 $\pm$ 6.42 (0.50) & 72.57 $\pm$ 6.79 (0.50)\\
 & Lymphoma & 69.50 $\pm$ 5.22 (0.00) & 64.00 $\pm$ 3.74 (0.00) & \textbf{71.50 $\pm$ 7.43} (0.50) & 65.00 $\pm$ 5.92 (0.50) & 65.00 $\pm$ 4.47 (0.50) & 61.50 $\pm$ 6.34 (0.50)\\
\midrule
Mass Spectrometry & Arcene & \textbf{66.50 $\pm$ 8.00} (0.00) & 66.00 $\pm$ 7.84 (0.00) & 62.75 $\pm$ 6.75 (0.25) & 62.00 $\pm$ 7.31 (0.25) & 54.25 $\pm$ 6.71 (0.80) & 60.75 $\pm$ 5.71 (0.80)\\
\midrule
Time-series & HAR & 94.35 $\pm$ 0.32 (0.00) & \textbf{94.72 $\pm$ 0.46} (0.00) & 94.29 $\pm$ 0.48 (0.25) & 94.70 $\pm$ 0.63 (0.25) & 94.62 $\pm$ 0.34 (0.50) & 94.72 $\pm$ 0.30 (0.50)\\      
\midrule
Speech & Isolet & 94.08 $\pm$ 0.45 (0.00) & 94.45 $\pm$ 0.56 (0.00) & \textbf{95.56 $\pm$ 0.30} (0.95) & 95.42 $\pm$ 0.42 (0.95) & 95.04 $\pm$ 0.37 (0.90) & 94.31 $\pm$ 1.27 (0.90)\\        
\midrule
Noisy & Madelon & \textbf{83.68 $\pm$ 1.89} (0.00) & 81.68 $\pm$ 3.00 (0.00) & 78.13 $\pm$ 2.41 (0.95) & 76.40 $\pm$ 2.15 (0.95) & 79.30 $\pm$ 1.61 (0.98) & 72.02 $\pm$ 2.94 (0.98)\\        
\midrule

            \end{tabular}
    }

 \end{table*}


\end{document}